\documentclass[10pt]{article}

\usepackage[preprint]{tmlr}
\usepackage[T1]{fontenc}

\usepackage{amsmath,amssymb,amsfonts,bm,mathtools}
\usepackage{amsthm}

\usepackage{graphicx}
\usepackage{subcaption}
\usepackage{booktabs}
\usepackage{threeparttable}
\usepackage{tabularx}
\usepackage{adjustbox}
\usepackage{pdflscape}
\usepackage{makecell}
\usepackage{multirow}

\usepackage[table]{xcolor}

\usepackage{float}
\usepackage{placeins}
\usepackage{wrapfig}

\usepackage{algorithm}
\usepackage{algpseudocode}

\usepackage{tikz}
\usetikzlibrary{calc,positioning}

\usepackage{microtype}
\usepackage{url}
\usepackage{hyperref}

\definecolor{NSPPKBlue}{RGB}{173,216,230}
\definecolor{passblue}{HTML}{DCEEFF}
\definecolor{failred}{HTML}{FFE5E5}
\definecolor{failgray}{HTML}{F0F0F0}

\tikzset{
  panel/.style={
    rounded corners=2mm,
    draw=black!15,
    fill=black!2,
    line width=0.4pt,
    inner sep=0pt
  }
}

\newlength{\pairheight}
\newcommand{\Subpanel}[5][]{%
  \begin{subfigure}[t]{0.45\linewidth}
    \centering
    \begin{tikzpicture}
      \node[panel] (P) {\includegraphics[width=\linewidth,#1]{#2}};
      \node[anchor=north west, font=\footnotesize\bfseries]
        at ([xshift=3pt,yshift=-3pt]P.north west) {(#3)};
    \end{tikzpicture}
    \caption{#4}
    \label{#5}
  \end{subfigure}
}

\newcommand{\TO}{\textbf{TIMEOUT}}

\newcommand{\mps}[2]{#1\,{\small$\pm$}\,#2}

\title{Structure Meets Attributes: A Sparse Graph Kernel Built from Neighborhood Pairs and Paths}

\author{
  \name Cosmina Simionescu \email cs843@exeter.ac.uk \\
  \addr Department of Computer Science \\
  University of Exeter
  \AND
  \name Fabrizio Costa \email f.costa@exeter.ac.uk \\
  \addr Department of Computer Science \\
  University of Exeter
}

\begin{document}

\maketitle

\begin{abstract}
Graph learning methods often require extensive tuning, while incorporating continuous node attributes into graph kernels can be computationally expensive. To tackle these issues we introduce the Neighborhood Subgraph Pairwise Path Kernel (NSPPK), which maps graphs with continuous node attributes into high-dimensional sparse feature vectors. NSPPK encodes graph structure through hashed pairs of rooted subgraphs augmented with shortest-path connectors, and incorporates continuous attributes through structure-dependent aggregation. This avoids attribute discretization and allows graph similarities to be computed as inner products between explicit feature vectors. Graph vectorization is deterministic, requires no training, and can be performed independently across graphs in parallel. With fixed structural parameters and bounded local adjacency sizes, its computational complexity is linear in the number of nodes. We examine the kernel’s discriminative properties and sensitivity to structural and attribute perturbations. Across six graph classification benchmarks, NSPPK achieves the best average accuracy rank among the evaluated graph kernels, without dataset-specific structural parameter tuning. Used as a vectorization method with XGBoost, it achieves higher mean classification accuracy than the evaluated neural baselines on five of the six datasets. Further experiments on QM9 and MolPCBA assess vectorization efficiency and applicability to larger molecular datasets. These results show that explicit sparse graph kernels can provide an effective alternative to learned graph embeddings for attributed graph classification.

\end{abstract}


\section{Introduction}

Graphs are a fundamental data structure for modeling relationships among
entities, with applications in social network analysis~\citep{newman2003structure},
protein function prediction~\citep{borgwardt2005protein}, molecular activity
prediction~\citep{kriege2012subgraph}, and recommender systems~\citep{ying2018graph}.
 Nodes can carry categorical labels,
such as atom types, and continuous attributes, such as spatial coordinates.
A central challenge in graph learning is to represent both graph structure
and node attributes in a form suitable for prediction and comparison.

Two widely used approaches to graph learning are graph kernels and graph
neural networks (GNNs). Graph kernels measure similarity between graphs,
typically by decomposing them into substructures and comparing their
occurrences~\citep{borgwardt2020graphkernels}. For example, the
Weisfeiler--Lehman subtree kernel constructs graph representations by
counting refined discrete node labels~\citep{shervashidze2011weisfeiler}.
In contrast, GNNs learn node representations by iteratively aggregating
and transforming information from neighboring nodes, which can then be
pooled to obtain graph-level representations~\citep{xu2019powerful}.
Both approaches are widely used in graph learning, but they differ in how graph representations are constructed and used.

For graph kernels, incorporating continuous node attributes alongside
structural information presents an additional challenge. Several methods
have been proposed to address this problem. Hash graph kernels adapt
discrete-label kernels to continuous attributes by repeatedly mapping
attribute values to discrete labels~\citep{morris2016fasterkernelsgraphscontinuous}.
GraphHopper compares node attributes at corresponding positions along
shortest paths~\citep{feragen2013scalable}. Propagation kernels diffuse
and quantize node attributes~\citep{neumann2016propagation}. Wasserstein
Weisfeiler--Lehman compares distributions of node representations by
solving a transport problem for each graph pair~\citep{togninalli2019wwl}.
These methods provide different ways of combining structural and
attribute information, but their feature construction and comparison
procedures can introduce additional computational costs.

Our goal is to construct a sparse, explicit feature representation
that integrates graph structure and continuous node attributes,
can be computed once per graph, and reused for graph similarity
comparisons and downstream learning tasks.

\begin{figure}[!htbp]
    \centering
    \hspace*{-1em}
    \includegraphics[width=0.75\textwidth, trim=0.3cm 2.4cm 12.3cm 2.0cm,  
    clip]{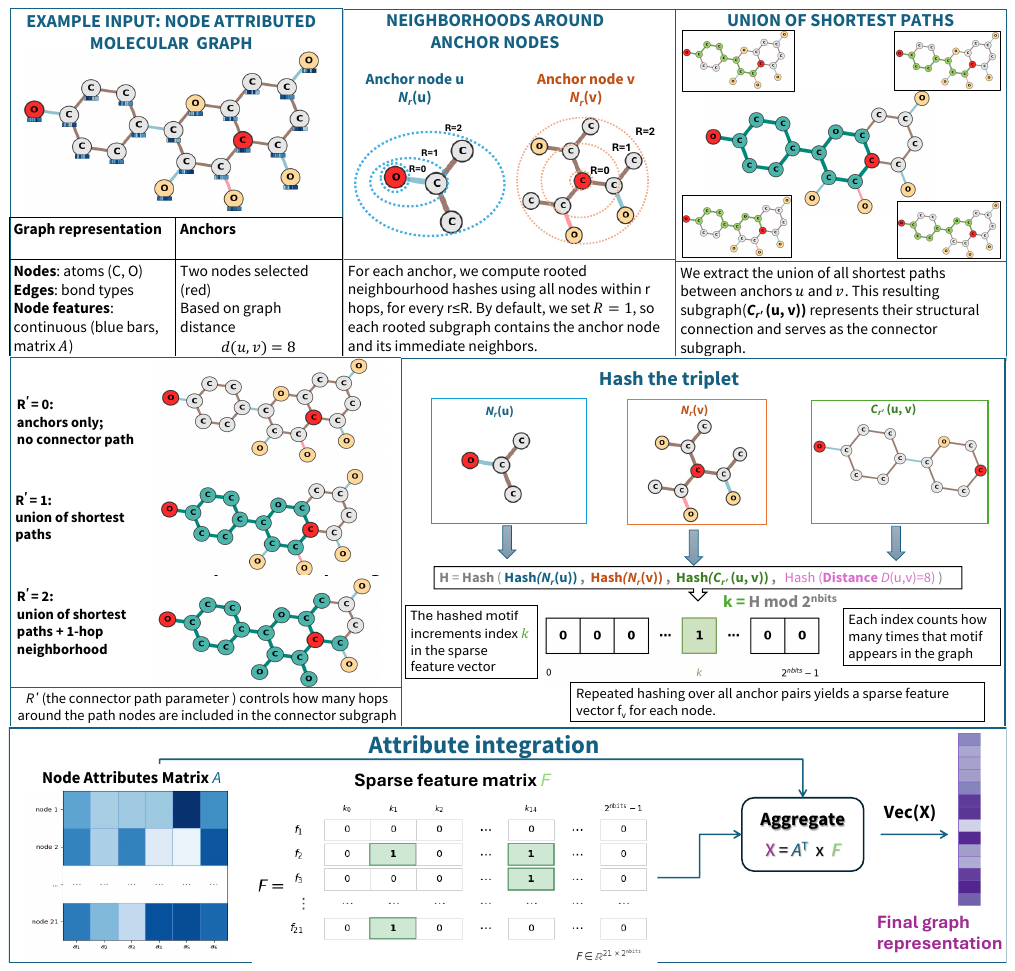}
    
\caption[Overview of NSPPK feature construction]{
NSPPK hashes the rooted neighborhoods of an unordered anchor pair $\{u,v\}$,
its connector, and its distance $d(u,v)$ to a bucket in a $2^b$-dimensional
vector. The three
connector examples on the left show, from top to bottom, the omitted connector,
the union of shortest paths, and the subgraph induced within one hop of that
union. Section~\ref{sec:notation} defines the connector notation.
Each occurrence increments the chosen bucket in both anchor rows of the sparse
matrix $M_G$ (labeled $F$ in the illustration). For node attributes $A$, the resulting graph vector
is $\phi_\theta(G,A)=\operatorname{vec}(A^\top M_G)$.}
    
\label{fig:nsppk_pipeline}
\end{figure}

We introduce the \emph{Neighborhood Subgraph Pairwise Path Kernel}
(NSPPK) for graphs with continuous node attributes. It builds on the
\emph{Neighborhood Subgraph Pairwise Distance Kernel}
(NSPDK)~\citep{costa2010fast}, which counts pairs of rooted
neighborhoods (subgraphs centered at individual nodes) using discrete
labels and the shortest-path distance between their roots, called
anchors. Matching neighborhoods and equal anchor distances can still leave
differences in the structure of the connecting paths. NSPPK incorporates
the union of all shortest paths between the anchors, with optional expansion
to nearby vertices, as part of the structural pattern. It hashes this pattern and efficiently integrates
the anchors' real-valued attribute vectors in the corresponding
feature bucket (Figure~\ref{fig:nsppk_pipeline}).

NSPPK maps each graph to a sparse feature vector, so graph similarity is computed as a dot product. Exact substructure matching can yield low similarity between distinct graphs~\citep{yanardag2015structural}: as neighborhoods grow, more of their structure must match exactly, and high-degree nodes can make even small-radius neighborhoods large. NSPPK decomposes graph structure into hashed features over pairs of rooted neighborhoods at multiple radii, augmented with connectors formed from the unions of shortest paths between their anchor nodes. This creates overlapping, path-based local patterns, allowing similarities to accumulate from shared structure without requiring an entire large neighborhood to match. Under fixed limits on neighborhood radius, anchor-pair distance, and connector expansion, feature extraction scales linearly in the number of nodes when local adjacency sizes are uniformly bounded (Section~\ref{sec:complexity}). We examine how NSPPK similarities respond to structural and attribute perturbations (Section~\ref{sec:Diagonal}).

\textbf{Empirical evaluation.}
We study graph similarity under perturbations, the runtime
of feature extraction, and the usefulness of the representation
for graph classification. With the same structural configuration
across the classification benchmarks, NSPPK attains the best
average rank among the evaluated graph kernels. With XGBoost,
its sparse features also rank first on average in the
comparison with graph neural network (GNN) baselines.

\section{Related Work}
\label{sec:relatedwork}
Many graph kernels are formulated within the \emph{R-convolution}
framework~\citep{haussler1999}, which decomposes structured objects into
parts and aggregates kernel evaluations over those parts. Examples include
kernels for parse trees~\citep{collins2001convolution}, string and tree
matching~\citep{vishwanathan2002fast}, and subpaths in unordered
trees~\citep{kimura2012fast}. \citet{cortes2004rational} show that Haussler's
convolution kernels for strings are special cases of rational kernels based
on weighted finite-state transducers. In graphs, the choice of
substructures determines both what the kernel can distinguish and the cost of
comparing graphs. Representative examples count
graphlets~\citep{shervashidze2009efficient} or record discrete node labels
refined by 1-WL~\citep{shervashidze2011weisfeiler}.

NSPDK~\citep{costa2010fast} counts occurrences of pairs of rooted
neighborhoods, using the neighborhood radius and the shortest-path distance
between the anchors to index the counts. The original experiments computed a
Gram matrix of pairwise graph similarities. Later implementations store the
counts in a sparse feature vector for each graph~\citep{navarin2017efficient}.
The original formulation uses discrete node and edge labels, so continuous
node attributes must first be encoded as discrete labels.
At a fixed radius, two anchor pairs with matching neighborhoods and equal
distances contribute to the same feature even if one pair has more shortest
paths than the other. NSPPK adds the union of those paths to the paired
neighborhoods and sums the anchors' real-valued attributes in each structural
bucket.

\citet{orsini2015graph} discuss NSPDK within their graph invariant kernel
framework. Their attributed neighborhood kernel uses structural similarity
between individual rooted neighborhoods to weight node attribute comparisons
across graphs. \citet{kriege2019explicit} derive approximate explicit
feature maps for GraphHopper and a graph invariant kernel with real-valued
attributes. These maps allow graph similarities to be computed as dot
products after feature extraction.

Random walk kernels compare label sequences along walks, and marginalized
kernels average over possible
walks~\citep{kashima2003marginalized,gartner2003graph,vishwanathan2010graph}.
These computations can be costly, and immediate edge reversals can yield
redundant walks, a problem known as \emph{tottering}~\citep{mahe2004extensions}.
Subgraph matching kernels compare attributes across compatible subgraph
correspondences. Computing the kernel involves enumerating cliques in a
product graph, so the cost grows with the number of admissible
matches~\citep{kriege2012subgraph}. Propagation kernels diffuse node attributes
and quantize the propagated values to build count
features~\citep{neumann2016propagation}. The learning-free \#WLS method
uses SimHash to sketch continuous node attributes and pools the sketches over
successive neighborhoods to embed each graph~\citep{wu2026learningfree}.
The shortest-path kernel compares
path lengths and endpoint information~\citep{borgwardt2005shortest}.
GraphHopper uses counts of occurrences at corresponding positions on
equal-length shortest paths to weight cross-graph node attribute comparisons.
Computing the kernel can therefore involve comparing attributes for all pairs of nodes across
the two graphs~\citep{feragen2013scalable}. Wasserstein WL compares distributions
of node representations, and fused Gromov--Wasserstein incorporates both
attributes and graph structure in the transport objective. Both solve a
transport problem for each graph pair~\citep{togninalli2019wwl,vayer2019optimaltransportstructureddata}.
Deep Graph Kernels learn latent representations of
substructures~\citep{yanardag2015deep}. Graph Neural Tangent Kernels derive
a kernel from the infinite-width limit of a chosen GNN
architecture~\citep{du2019graphnt}.

Several recent kernels compare distributions of node or substructure
representations. MMD-GK compares distributions of propagated node
representations using maximum mean discrepancy. Direct evaluation of the
discrepancy for a graph pair has a term quadratic in the node count, separate
from feature propagation~\citep[Theorem~3]{mmdgk_iclr2024}. Sliced
Wasserstein Weisfeiler--Lehman (SWWL) represents each graph by quantiles of
random projections of its propagated node features. The numbers of projections
and quantiles determine the size of the feature vector and the resolution of
the approximation~\citep{carpintero2024swwl}. The Distributional Shortest-Path
(DSP) kernel learns node representations from shortest paths with a neural
language model. It constructs an explicit graph vector by averaging
partition-kernel features of the learned node representations using
partitions formed across the dataset~\citep[Section~IV-B]{ye2025dsp}.

In NSPPK, the two rooted neighborhoods, their anchor distance, and the
shortest-path connector jointly determine a structural bucket. Each bucket
stores the sum of the real-valued attributes of its anchors, computed
independently for each graph. The resulting sparse vectors can be compared
by a dot product in time linear in their combined number of nonzero entries
for sorted sparse indices (Section~\ref{sec:complexity}). Different sets of
node attributes with the same sum contribute the same bucket values.

\section{Method}
\label{sec:method}

An R-convolution kernel decomposes a structured object and sums kernel
evaluations over its parts~\citep{haussler1999}. NSPDK decomposes a graph into
pairs of equal-radius rooted neighborhoods whose anchors are at a prescribed
shortest-path distance~\citep{costa2010fast}. NSPPK can augment each pair with a
connector. All shortest paths between the anchors contribute their vertices
and edges to a single subgraph, which can be expanded to include nearby
vertices. After hashing the structural pattern, NSPPK adds the real-valued
attribute vectors of the two anchors to the corresponding feature bucket.

\subsection{Core Objects and Notation}
\label{sec:notation}

We assume standard graph-theoretic notation; for completeness, basic definitions
(graph, degree, shortest-path distance, $r$-hop neighborhoods) are collected in
Appendix~\ref{sec:definitions}.
Here we introduce only the concepts that are specific to NSPPK.
We write $R$ for the maximum anchor-neighborhood radius, $D$ for the maximum
anchor-pair distance, and $R'$ for the maximum connector radius, with
$R,R'\in\mathbb{N}_0$ and $D\ge 1$.
We write $K_{\max}=\max_{v\in V}\deg(v)$ for the maximum degree and $\tau$ for
an optional degree cutoff.
We refer to $(R,D,R')$ as the core structural hyperparameters of NSPPK; the
hash width $b$ and degree cutoff $\tau$ are implementation controls for the
collision--memory and feature-size trade-offs.

\textbf{Anchor pairs.}
We consider (unordered) anchor pairs $\{u,v\}\subseteq V$ with $u\neq v$ and
finite shortest-path distance $d(u,v)\le D$; disconnected pairs are ignored.

\textbf{Union of shortest paths.}
Let $U(u,v)$ contain exactly the vertices and edges that lie on at least
one shortest path between $u$ and $v$.

\textbf{Connector (thickened path union).}
For a vertex set $S\subseteq V$, let $N_{r'}(S)$ denote the induced subgraph on
all vertices within shortest-path distance at most $r'$ from $S$, i.e.,
$N_{r'}(S)=G\!\left[\{x\in V:\min_{s\in S} d(x,s)\le r'\}\right]$.
Write $C_0(u,v)=U(u,v)$. For $r'>0$, set
$C_{r'}(u,v)=N_{r'}(V(U(u,v)))$, the subgraph induced by vertices within
$r'$ hops of the path union.

\textbf{NSPPK feature primitives.}
An NSPPK feature is defined by a pair of anchor neighborhoods
$(N_{r_u}(u),N_{r_v}(v))$ whose centers are at shortest-path distance $d(u,v)$,
optionally augmented with a connector subgraph $C_{r'}(u,v)$. Features are
parameterized by the tuple $(r_u,r_v,d,r')$ and are invariant to permutations of
vertices within each subgraph.

\subsection{Augmented structural features}
For each anchor pair, NSPPK enumerates endpoint radii
$r_u,r_v\in\{0,\ldots,R\}$ and connector radii
$r'\in\{0,\ldots,R'\}\cup\{\varnothing\}$. The configuration includes the
observed distance $d=d(u,v)\in\{1,\ldots,D\}$. The sentinel
$r'=\varnothing$ gives an endpoint-only feature.

\subsection{From subgraphs to explicit features}

NSPPK constructs an explicit feature representation for each
graph independently by enumerating structural patterns associated
with pairs of nodes. Each pattern comprises their rooted
neighborhoods at specified radii, their shortest-path distance,
and an optional shortest-path connector.

Each pattern is mapped to a deterministic feature bucket using
permutation-invariant hashing. Let $b$ denote the hash width
and $B=2^b$ the number of available feature buckets. Since the
hash space is finite, distinct patterns may collide; we
distinguish these practical bucket indices from the idealized
collision-free identifiers used in the analysis.

Let $\theta=(R,D,R')$ denote the structural parameters, with
$b$ and the optional degree cutoff $\tau$ treated as fixed
implementation settings.

For a graph $G=(V,E)$ without continuous node attributes,
the feature map
$\phi_\theta(G)\in\mathbb{N}_0^B$ records the number of
structural pattern occurrences in each bucket, counting each
occurrence once per unordered node pair and structural
configuration. The corresponding kernel is
\[
\kappa_\theta(G,G')
=
\phi_\theta(G)^\top\phi_\theta(G').
\]

For a node-attributed graph, let
$a:V\to\mathbb{R}^p$ be the node attribute function and
$A\in\mathbb{R}^{|V|\times p}$ its matrix representation,
with row $A_{v,:}=a(v)^\top$.Rather than counting structural pattern occurrences, each
feature bucket aggregates the attribute vectors of the two
nodes associated with each occurrence. This defines the
attributed feature map
$\phi_\theta(G,A)\in\mathbb{R}^{pB}$, detailed in
Section~\ref{subsec:attribute_integration}.

For two attributed graphs $(G,A)$ and $(G',A')$ with the
same attribute dimension and feature-extraction settings,
the kernel is
\[
\kappa_\theta\bigl((G,A),(G',A')\bigr)
=
\phi_\theta(G,A)^\top\phi_\theta(G',A').
\]

Both kernels are positive semidefinite by construction,
as they are inner products between explicit feature vectors.
\subsection{Feature Hashing Pipeline}
\label{subsec:feature_hashing}
NSPPK maps each subgraph pattern to a deterministic integer bucket in
$\{0,\dots,2^b-1\}$ using permutation-invariant hash functions, where $b$ is
the number of hash bits and $B=2^b$ is the number of hash buckets. This mapping
is not injective for finite $b$. 
Hashes are computed \emph{incrementally} with respect to neighborhood radius
and graph distance, and are \emph{cached per node}. As a result, all rooted
neighborhood hashes are computed once and reused when forming neighborhood
pairs and connector features, avoiding quadratic recomputation over node pairs
of the neighborhood hashes. The conditions under which total feature extraction
is near-linear are stated in Section~\ref{sec:complexity}.

The resulting bucket index provides constant-time indexing into the feature
vector and avoids explicit subgraph isomorphism checks.

\textbf{Base hash functions.}
For any element $x$, $h_b(x) = \mathrm{sha256}(x) \bmod 2^b$ is the $b$-bit base
hash. From $h_b$ we define:
$h_{\mathrm{seq}}(I)$: \emph{sequence hash} of an ordered tuple; $I = (x_1,\dots,x_s)$;
$h_{\mathrm{multi}}(S)$: \emph{multiset hash} of $S$, computed after lexicographic sorting
to ensure order invariance. Both compound hash functions return indices in
$\{0,\ldots,2^b-1\}$. Collections passed to $h_{\mathrm{multi}}$ retain repeated
items.

\textbf{Node hash.}
For each node $v$ (labels and neighborhoods as in
Section~\ref{sec:definitions}), we define
$\eta(v)=h_{\mathrm{seq}}([h_b(\ell(v)),h_{\mathrm{multi}}(\{\,h_{\mathrm{seq}}([h_b(\ell(u)),h_b(\ell(\{v,u\}))]) : u\in N(v)\,\})])$.

\textbf{Rooted graph hash.}
For radius $r$, set $\sigma_j^v = h_{\mathrm{multi}}(\{\, \eta(u) : u \in \mathcal{S}_j(v) \,\})$ with
$\mathcal{S}_j(v) = \{\,u \mid d(v,u) = j \,\}$ and
$\zeta_r(v) = h_{\mathrm{seq}}([\sigma_0^v,\, \sigma_1^v,\, \dots,\, \sigma_r^v])$.
These hashes are computed incrementally in $r$ and cached for each node.

\textbf{Endpoint-pair code.}
Given cached rooted hashes, define the endpoint descriptor
$E_r(x)=h_{\mathrm{seq}}([h_b(r),\zeta_r(x)])$. For nodes $u,v$ at distance
$d$, the unordered endpoint-pair code is
\[
P_{\mathrm{pair}}^{r_u,r_v}(u,v)
=h_{\mathrm{multi}}(\{E_{r_u}(u),E_{r_v}(v)\}).
\]
Including the radius in each endpoint descriptor makes this code invariant to
exchanging $(u,r_u)$ and $(v,r_v)$, including when $r_u\ne r_v$.

\textbf{Connector code.}
For $r'\geq 0$, let $C=C_{r'}(u,v)$ and $H=\max(R,R')$. For an anchor
$x\in\{u,v\}$, form the shell signature
\[
S_x(C)=h_{\mathrm{seq}}\!\left(
  \left[
  h_{\mathrm{multi}}\!\left(
    \{\zeta_H(w):w\in V(C),\ d_C(x,w)=j\}
  \right)
  \right]_{j=0}^{d+r'}
\right),
\]
where $d_C$ is shortest-path distance within the connector. Thus the
multisets are explicitly restricted to vertices of the thickened connector,
rather than all vertices in $G$ at distance $j$ from an anchor. The connector
code is
\[
Q_{r'}(u,v)=h_{\mathrm{seq}}\!\left(
 [h_b(r'),h_{\mathrm{multi}}(\{S_u(C),S_v(C)\})]
\right).
\]
The multiset over the two anchor views makes the code invariant to exchanging
$u$ and $v$, while $h_b(r')$ records the thickening radius. We use a reserved
sentinel $Q_{\varnothing}$ when the connector is omitted.

\textbf{Joint feature and final vector.}
The bucket associated with one NSPPK configuration is the combined hash
\[
F_{r_u,r_v,d,r'}(u,v)=h_{\mathrm{seq}}\!\left([
 h_b(d),P_{\mathrm{pair}}^{r_u,r_v}(u,v),
 \begin{cases}
 Q_{r'}(u,v), & r'\geq 0,\\
 Q_{\varnothing}, & r'=\varnothing
 \end{cases}
 ]\right).
\]
The NSPPK vector $\phi_\theta(G)$ is the histogram of these joint bucket values
over all unordered anchor pairs and all enumerated $(r_u,r_v,d,r')$
configurations. Pair and connector codes are therefore components of one
feature, not independent histogram entries.


\subsection{Node attribute integration}
\label{subsec:attribute_integration}
Because NSPPK features are extracted locally around anchor pairs, each node is
incident to a well-defined subset of feature occurrences: those for which it is
one of the two anchors. Continuous node attributes are integrated using this
endpoint incidence; internal vertices of an endpoint neighborhood or connector
do not contribute their attributes to that occurrence.
Formally, for a graph with continuous node attributes
$A\in\mathbb{R}^{|V|\times p}$, let
$M_G\in\mathbb{N}_0^{|V|\times B}$ denote the node-feature count matrix,
where $B=2^b$. Its entries are defined operationally by
\[
(M_G)_{xj}=
\sum_{\substack{\{u,v\},\,r_u,r_v,d,r'\\
F_{r_u,r_v,d,r'}(u,v)=j}}
\mathbf{1}\{x\in\{u,v\}\}.
\]
Thus an occurrence increments the row of each of its two distinct anchors,
including repeated occurrences in the same bucket. We define the attributed
NSPPK feature vector as
\[
\phi_\theta(G,A)=\operatorname{vec}(A^\top M_G)\in\mathbb{R}^{pB}.
\]
When attributes are not used, we write $\phi_\theta(G)\in\mathbb{N}_0^B$
for the corresponding structural feature histogram. Thus each feature index in
the attributed vector stores the sum of attribute vectors of anchor endpoints
of occurrences mapped to that bucket. Node weights can be
incorporated by replacing $A$ with $\mathrm{diag}(w)A$, and an analogous edge-feature incidence matrix would be needed for edge attributes.

The resulting attributed kernel is positive semidefinite because it is the
ordinary inner product between explicit finite-dimensional vectors
$\phi_\theta(G,A)$. It is not invariant to arbitrary rescaling of the raw
attributes: as with any linear kernel on continuous inputs, attribute dimensions
with larger numerical scale can receive larger weight. In practice, attributes
should therefore be standardized or otherwise normalized when their units are
not directly comparable. The experiments below evaluate node attributes; edge
attributes can be included by applying the same aggregation principle to
edge-feature incidence matrices, but they are not the focus of the reported
benchmarks.

We emphasize that NSPPK feature extraction is entirely deterministic and
training-free; supervised learning is only performed by the downstream
classifier.





\subsection{Complexity Analysis}
\label{sec:complexity}

Let $\mathcal{B}_\ell(v)$ be the $\ell$-hop ball of $v$ and
$\mathcal{E}_\ell(v)$ the edges incident to that ball. With $H=\max(R,R')$,
the truncated traversals for the rooted hashes and anchor enumeration cost
\[
O\!\left(
\sum_{v\in V} (|\mathcal{B}_H(v)|+|\mathcal{E}_H(v)|)
+
\sum_{v\in V} (|\mathcal{B}_D(v)|+|\mathcal{E}_D(v)|)
\right).
\]
Feature extraction also pays for sorting hash multisets, enumerating radius
combinations, and constructing and hashing the connectors. Appendix~\ref{sec:appendix:complexity}
accounts for these costs, including implementations that list all shortest
paths. With fixed radii and uniformly bounded local adjacency sizes, each
node and each eligible anchor pair contributes bounded structural work, so
feature extraction scales linearly in $|V|$ under those assumptions.

Sparse aggregation of $p$ attribute channels takes
$O(p\,\operatorname{nnz}(M_G))$ operations, plus any dense-output
initialization. For two vectors $x$ and $y$ stored with sorted nonzero indices,
the kernel dot product takes
$O(\operatorname{nnz}(x)+\operatorname{nnz}(y))$ time. This applies to the
attributed kernel with $x=\phi_\theta(G,A)$ and $y=\phi_\theta(G',A')$.

\subsection{Balancing Feature Size and Hash Collisions}
\begin{figure}[t]
  \centering
  \begin{subfigure}[t]{0.46\linewidth}
    \centering
    \includegraphics[
      width=\linewidth,
      keepaspectratio,
      trim=6 6 6 6,
      clip
    ]{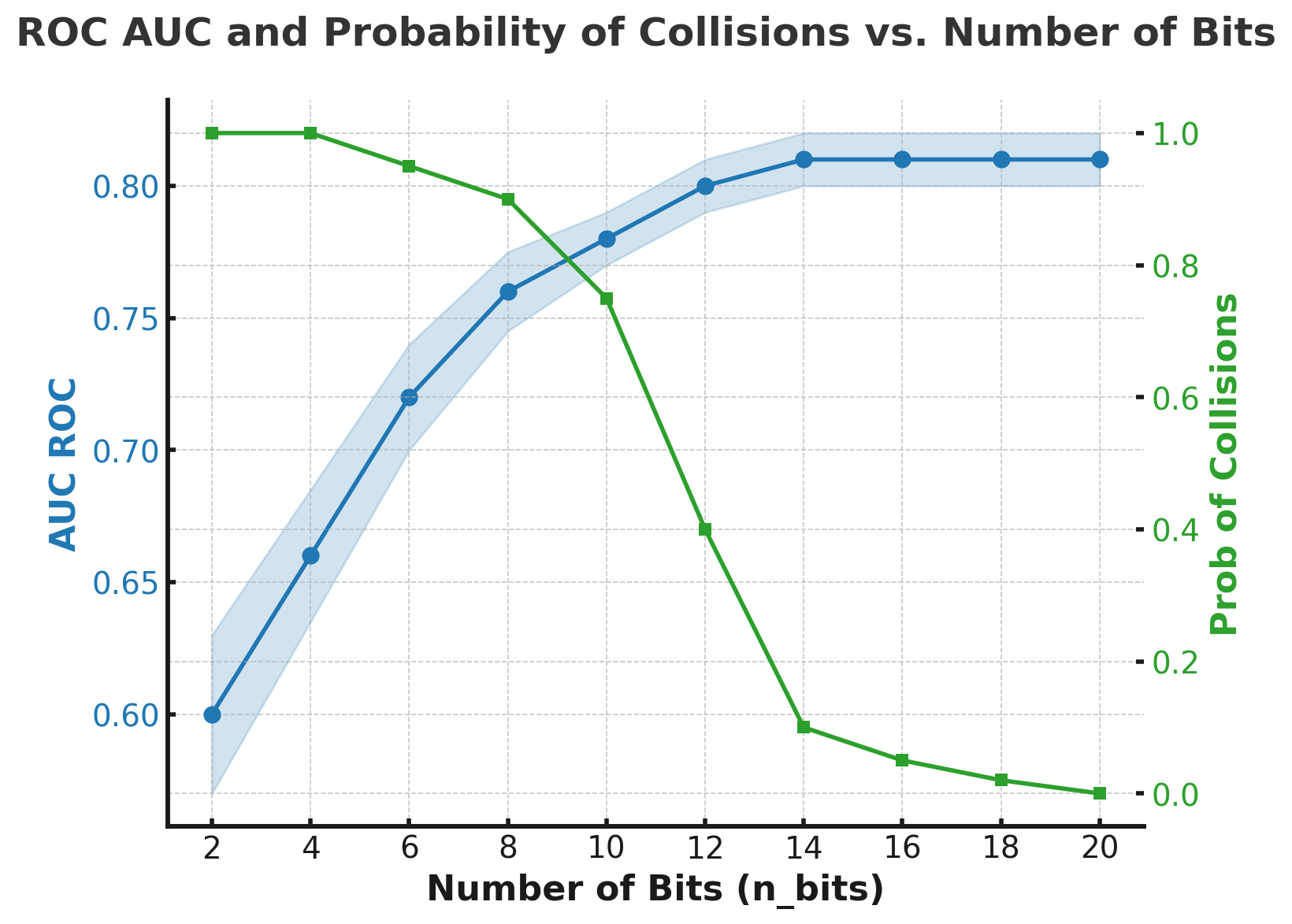}
    \caption{Test ROC--AUC and collision probability by hash width.}
    \label{fig:nbit}
  \end{subfigure}
  \hfill
  \begin{subfigure}[t]{0.50\linewidth}
    \centering
    \includegraphics[
      width=\linewidth,
      keepaspectratio
    ]{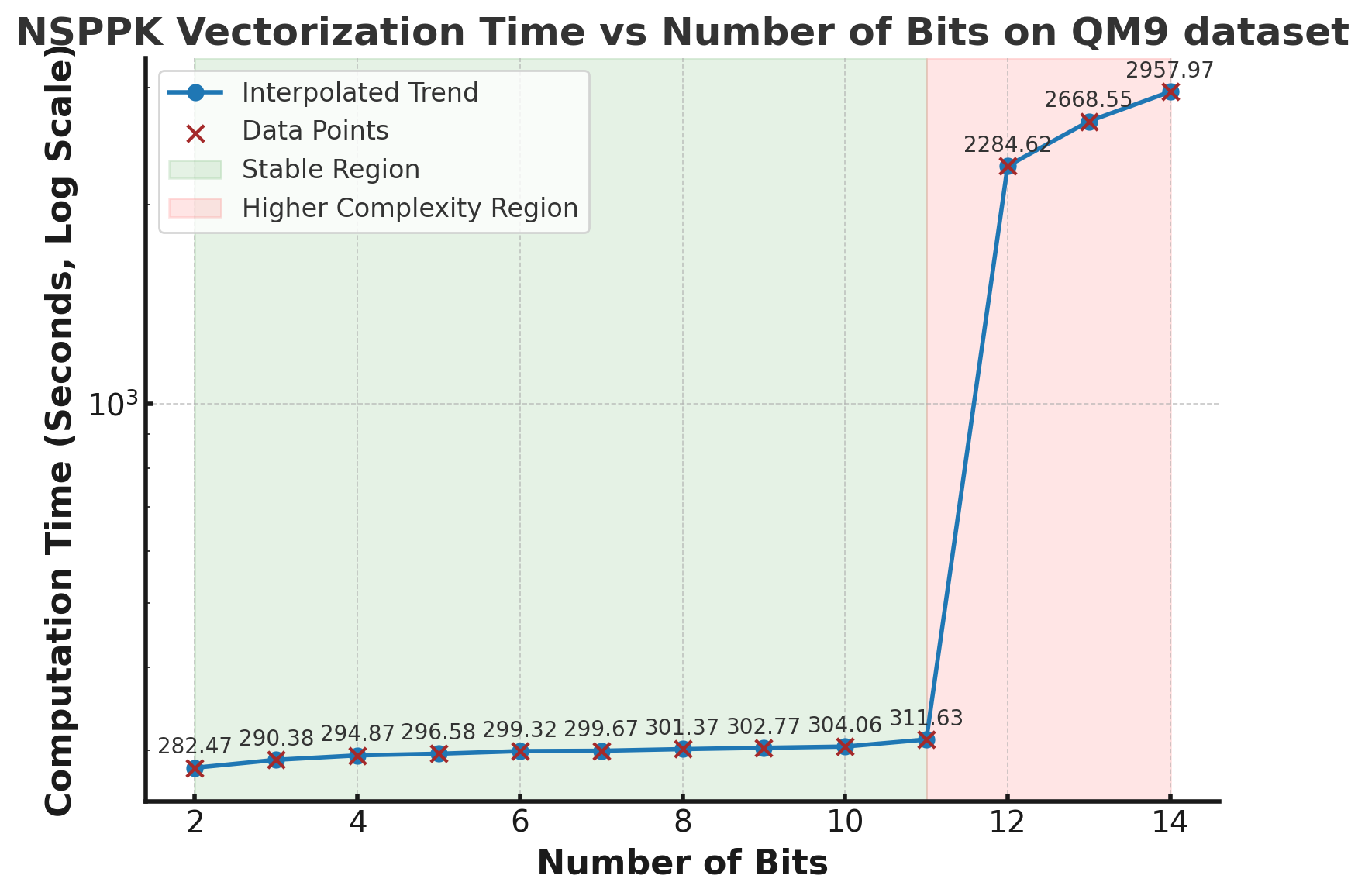}
    \caption{NSPPK vectorization time on QM9 vs.\ number of hash bits.}
    \label{fig:nsppk_qm9_main}
  \end{subfigure}
  \caption{Hash-width sensitivity on two molecular graph collections. The left
  panel reports the PubChem classification experiment. The right panel reports
  QM9 vectorization time.}
  \label{fig:bits_side_by_side}
\end{figure}
The encoding allocates $2^b$ feature coordinates at hash width $b$. We measured test ROC--AUC for a
random forest trained on 1{,}300 molecular graphs from PubChem AID~463230,
a plasma platelet-activating factor acetylhydrolase inhibition
assay~\citep{pubchem463230}. The experiment omits continuous node attributes.
Figure~\ref{fig:bits_side_by_side}(a) shows ROC--AUC rising from about $0.60$
at 2 bits to $0.81$ at 14 bits and remaining at about $0.81$ through 20 bits.
The plotted collision probability is about $0.10$ at 14 bits.
On the preprocessed QM9 collection, vectorization time rose from 312 seconds
at 11 bits to 2{,}285 seconds at 12 bits in the reported environment
(Appendix~\ref{sec:qm9}).

\section{Isomorphism Invariance and Structural Discrimination}
\label{sec:isomorphism}

A fundamental property of graph kernels is invariance under
graph isomorphism. Two graphs $G=(V_G,E_G)$ and $G'=(V_{G'},E_{G'})$
are isomorphic, denoted $G\cong G'$, if there exists a bijection
$f:V_G\to V_{G'}$ that preserves adjacency and, for labeled
graphs, node and edge labels.

An isomorphism-invariant feature map satisfies
\[
G\cong G'
\quad\Longrightarrow\quad
\phi_\theta(G)=\phi_\theta(G').
\]
The converse does not generally hold: non-isomorphic graphs
may be mapped to identical feature vectors. The discriminative
power of a graph representation therefore depends on the
structural differences it can distinguish.

NSPPK is isomorphism-invariant by construction. Rooted
neighborhood descriptors, shortest-path distances, and
connector signatures are preserved under graph isomorphism.
Their permutation-invariant hashing and aggregation ensure
that the resulting feature vector is independent of vertex
ordering.

We examine the structural discrimination of NSPPK relative
to the one-dimensional Weisfeiler--Lehman (1-WL) test and
bounded-radius NSPDK. In the following examples, node and
edge labels are constant, continuous node attributes are
omitted, and structural feature identifiers are treated as
collision-free.

\paragraph{Comparison with message passing.}

The discriminative power of many graph kernels and graph
neural networks is related to the 1-WL test
\citep{shervashidze2011weisfeiler,morris2019weisfeiler}.
In particular, standard message-passing GNNs are at most
as discriminative as 1-WL~\citep{xu2019powerful}, while the
WL subtree kernel constructs graph representations from
the colors generated during its iterative refinement.

\begin{figure}[H]
\vspace{-0.4em}
\centering
\begin{adjustbox}{max width=0.9\linewidth}
\begin{tikzpicture}[x=1cm,y=1cm,
                    line cap=round, line join=round,
                    every node/.style={outer sep=0pt}]

  \tikzset{
    panel/.style={
      rounded corners=2mm,
      draw=black!20,
      fill=black!2,
      line width=0.4pt
    },
    edge/.style={
      line width=0.95pt,
      draw=black!70
    },
    vblue/.style={
      circle, draw=blue!55!black, fill=blue!40,
      minimum size=6.2pt, inner sep=0pt
    },
    vgreen/.style={
      circle, draw=teal!55!black, fill=teal!40,
      minimum size=6.2pt, inner sep=0pt
    },
    vred/.style={
      circle, draw=red!55!black, fill=red!40,
      minimum size=6.2pt, inner sep=0pt
    },
    vyellow/.style={
      circle, draw=orange!55!black, fill=orange!40,
      minimum size=6.2pt, inner sep=0pt
    }
  }

  \node[panel, minimum width=6.6cm,
        minimum height=2.7cm, anchor=center]
        (A) at (0,0) {};

  \node[panel, minimum width=6.6cm,
        minimum height=2.7cm, anchor=center]
        (B) at (7.6,0) {};

  \node[anchor=west]
    at ([xshift=-3.05cm,yshift=1.05cm]A.center)
    {\small\textbf{(a)}};

  \node[anchor=west]
    at ([xshift=-3.05cm,yshift=1.05cm]B.center)
    {\small\textbf{(b)}};

  \path[use as bounding box]
    ([xshift=-1mm,yshift=-1mm]A.south west)
    rectangle
    ([xshift=+1mm,yshift=+1mm]B.north east);

  \begin{scope}
    \clip[rounded corners=2mm]
      (A.south west) rectangle (A.north east);

    \def\R{0.85}

    \foreach \i in {0,...,5}
      \coordinate (h\i) at
      ($([xshift=-1.8cm]A.center)+(\i*60:\R)$);

    \foreach \i/\j in {0/1,1/2,2/3,3/4,4/5,5/0}
      \draw[edge] (h\i)--(h\j);

    \foreach \i in {0,...,5}
      \node[vblue] at (h\i) {};

    \coordinate (S) at
      ($([xshift=+1.65cm]A.center)$);

    \coordinate (u1) at ($(S)+(-0.95,  0.45)$);
    \coordinate (u2) at ($(S)+(-0.35,  0.45)$);
    \coordinate (u3) at ($(S)+(-0.65, -0.35)$);
    \coordinate (v1) at ($(S)+( 0.35, -0.45)$);
    \coordinate (v2) at ($(S)+( 0.95, -0.45)$);
    \coordinate (v3) at ($(S)+( 0.65,  0.35)$);

    \foreach \a/\b in
      {u1/u2,u2/u3,u3/u1,v1/v2,v2/v3,v3/v1}
      \draw[edge] (\a)--(\b);

    \foreach \p in {u1,u2,u3,v1,v2,v3}
      \node[vgreen] at (\p) {};
  \end{scope}

  \begin{scope}
    \clip[rounded corners=2mm]
      (B.south west) rectangle (B.north east);

    \coordinate (c) at
      ($([xshift=-1.8cm]B.center)+(0,0)$);

    \coordinate (l1) at
      ($([xshift=-1.8cm]B.center)+(-0.7,0.55)$);

    \coordinate (l2) at
      ($([xshift=-1.8cm]B.center)+(-0.95,-0.5)$);

    \coordinate (r1) at
      ($([xshift=-1.8cm]B.center)+(0.7,0.55)$);

    \coordinate (r2) at
      ($([xshift=-1.8cm]B.center)+(0.95,-0.5)$);

    \foreach \a/\b in
      {l1/c,c/l2,l2/l1,r1/c,c/r2,r2/r1}
      \draw[edge] (\a)--(\b);

    \foreach \p in {l1,l2,c,r1,r2}
      \node[vred] at (\p) {};

    \coordinate (g00) at
      ($([xshift=+1.7cm]B.center)+(-0.95,0.5)$);

    \coordinate (g01) at
      ($([xshift=+1.7cm]B.center)+(-0.95,-0.5)$);

    \coordinate (g10) at
      ($([xshift=+1.7cm]B.center)+(0.0,0.5)$);

    \coordinate (g11) at
      ($([xshift=+1.7cm]B.center)+(0.0,-0.5)$);

    \coordinate (g20) at
      ($([xshift=+1.7cm]B.center)+(0.95,0.5)$);

    \coordinate (g21) at
      ($([xshift=+1.7cm]B.center)+(0.95,-0.5)$);

    \foreach \a/\b in
      {g00/g10,g10/g20,g01/g11,g11/g21}
      \draw[edge] (\a)--(\b);

    \foreach \a/\b in
      {g00/g01,g10/g11,g20/g21}
      \draw[edge] (\a)--(\b);

    \foreach \p in {g00,g01,g10,g11,g20,g21}
      \node[vyellow] at (\p) {};
  \end{scope}

\end{tikzpicture}
\end{adjustbox}

\caption{Structural discrimination by NSPPK and 1-WL.
\textbf{(a)} A six-cycle $C_6$ and two disjoint triangles
$K_3\sqcup K_3$, distinguished by NSPPK for $D\geq 2$
but not by 1-WL.
\textbf{(b)} A bow-tie graph and a $2\times3$ ladder,
distinguished by both methods.
Colors are used solely to differentiate the drawings;
node and edge labels are constant in the comparisons.}
\label{fig:isomorphism}
\end{figure}
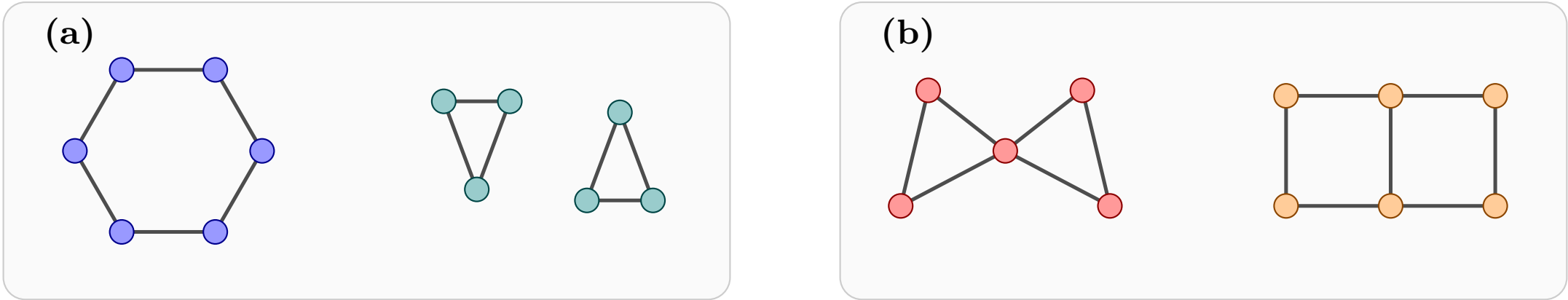

Figure~\ref{fig:isomorphism}(a) illustrates two non-isomorphic
graphs that are indistinguishable by 1-WL: a six-cycle
$C_6$ and two disjoint triangles $K_3\sqcup K_3$.
Both graphs contain six vertices of degree two, resulting
in identical color histograms at every refinement iteration.

NSPPK distinguishes these graphs for $D\geq 2$, since
pairs of nodes at shortest-path distance two occur in
the six-cycle but not in the two disjoint triangles.
This distinction follows from the inclusion of pairwise
distances in the structural representation and does not
require connector features.

Figure~\ref{fig:isomorphism}(b) illustrates a case in
which both 1-WL and NSPPK distinguish the graphs.
The bow-tie graph and the ladder have different degree
distributions and different numbers of node pairs at
distance one.

\paragraph{Comparison with bounded-radius NSPDK.}

Although shortest-path distances provide information about
the separation between nodes, they do not characterize the
structure of the paths connecting them. Consequently, node
pairs with identical rooted neighborhoods and equal distances
may contribute identical NSPDK features despite differences
in their connecting subgraphs.

NSPPK extends this representation by incorporating
shortest-path connectors into the structural feature
encoding. We illustrate the additional discriminative
capacity of these connectors through a pair of graphs
that cannot be distinguished by bounded-radius NSPDK.

Let $\phi^{\mathrm{NSPDK}}_{R,D}$ denote the NSPDK feature
map with maximum neighborhood radius $R$ and maximum
pairwise distance $D$~\citep{costa2010fast}.

\begin{figure}[H]
  \centering
  \begin{adjustbox}{width=0.64\linewidth}
  \begin{tikzpicture}[
      line cap=round,
      line join=round,
      edge/.style={
        draw=black!55,
        line width=0.75pt
      },
      path edge/.style={
        draw=blue!65!black,
        line width=1.2pt
      },
      vertex/.style={
        circle,
        draw=black!65,
        fill=white,
        minimum size=4.6pt,
        inner sep=0pt
      },
      anchor vertex/.style={
        vertex,
        draw=blue!65!black,
        fill=blue!65!black
      }]

    \node[anchor=west,font=\small\bfseries]
      at (-1.65,0.97) {(a)};

    \node[anchor=west,font=\small\bfseries]
      at (2.28,0.97) {(b)};

    \foreach \i/\ang in
      {0/90,1/18,2/306,3/234,4/162} {
      \coordinate (p\i) at (\ang:1.11cm);
      \coordinate (q\i) at (\ang:0.56cm);
    }

    \foreach \i/\j in {0/1,1/2,2/3,3/4,4/0} {
      \draw[edge]
        (p\i)--(p\j)
        (q\i)--(q\j);
    }

    \foreach \i in {0,...,4} {
      \draw[edge] (p\i)--(q\i);
    }

    \draw[path edge]
      (p0)--(p1)--(q1)--(q0)--(p0);

    \foreach \i in {0,...,4} {
      \node[vertex] at (p\i) {};
      \node[vertex] at (q\i) {};
    }

    \node[anchor vertex] at (p0) {};
    \node[anchor vertex] at (q1) {};

    \coordinate (a0) at (2.85,0.49);
    \coordinate (a1) at (2.85,-0.49);
    \coordinate (b0) at (6.15,0.49);
    \coordinate (b1) at (6.15,-0.49);

    \foreach \i/\y in {2/0.8,3/0,4/-0.8} {
      \coordinate (a\i) at (4.98,\y);
      \coordinate (b\i) at (4.02,\y);

      \draw[edge]
        (b\i)--(a\i)
        (a\i)--(b0)
        (a\i)--(b1);

      \draw[path edge]
        (a0)--(b\i)--(a1);
    }

    \foreach \i in {0,...,4} {
      \node[vertex] at (a\i) {};
      \node[vertex] at (b\i) {};
    }

    \node[anchor vertex] at (a0) {};
    \node[anchor vertex] at (a1) {};

  \end{tikzpicture}
  \end{adjustbox}

  \caption{Connector-based discrimination beyond
  bounded-radius NSPDK.
  \textbf{(a)} Pentagonal prism $G$.
  \textbf{(b)} Graph $G'$ obtained by deleting disjoint
  four- and six-cycles from $K_{5,5}$.
  Both graphs have identical NSPDK representations for
  $R=1$ and $D=2$.
  Filled vertices indicate selected node pairs at distance
  two, and blue edges indicate their shortest-path unions
  $C_0(u,v)$, containing two shortest paths in (a)
  and three in (b).}
  \label{fig:connector_refinement}
\end{figure}
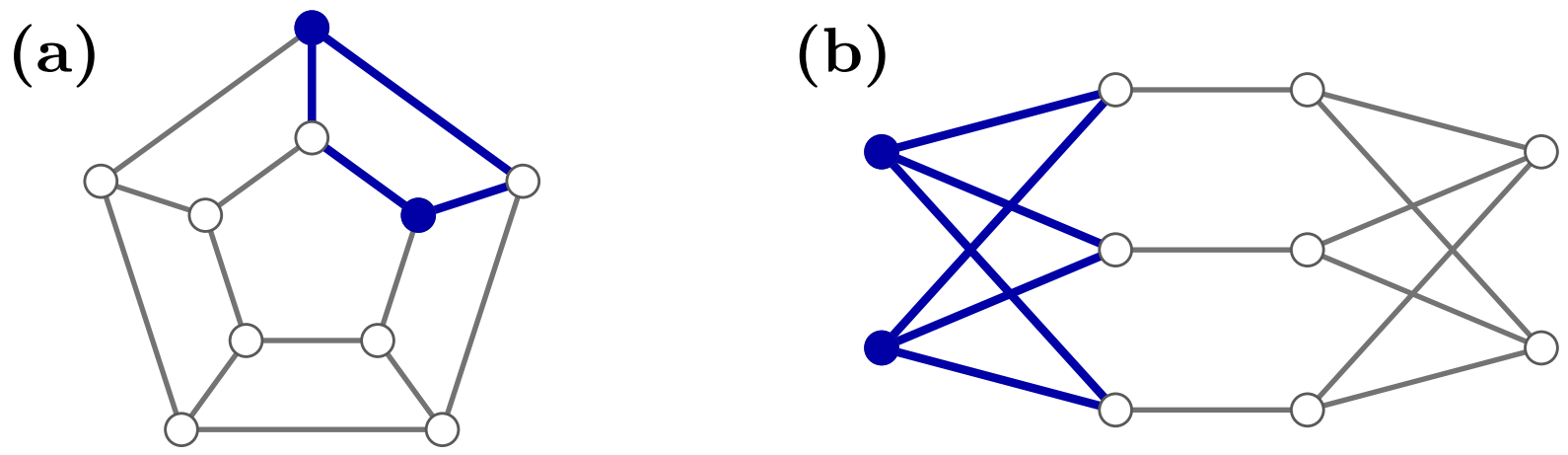

Consider the two graphs in
Figure~\ref{fig:connector_refinement}. Graph $G$ is a
pentagonal prism, while $G'$ is obtained from $K_{5,5}$
by deleting a four-cycle and a vertex-disjoint six-cycle.Both graphs are 3-regular and triangle-free, with ten
vertices and fifteen edges. They therefore have identical
rooted neighborhood types at radii zero and one.
Furthermore, each graph contains fifteen unordered node
pairs at distance one and twenty at distance two.

For $R=1$ and $D=2$, these properties result in identical
bounded-radius NSPDK representations.

Despite their identical bounded-radius NSPDK representations,
the two graphs differ in the organization of their shortest
paths. In $G$, any pair of nodes at distance two has at most
two common neighbors. In $G'$, the two vertices in each
bipartition that belonged to the deleted four-cycle share
three common neighbors.

With $R'=0$, NSPPK includes the union of all shortest
paths between each eligible pair of nodes. For the selected
pairs in Figure~\ref{fig:connector_refinement}, the first
distance shell within the connector contains two vertices
in $G$ and three in $G'$.

The connector signature defined in
Section~\ref{subsec:feature_hashing} preserves this
multiplicity through its multiset construction.
Consequently, at $r_u=r_v=1$, a joint structural feature
occurs in $G'$ that does not occur in $G$.

Under collision-free structural encoding, the two
feature maps therefore satisfy
\[
\phi^{\mathrm{NSPDK}}_{1,2}(G)
=
\phi^{\mathrm{NSPDK}}_{1,2}(G'),
\qquad
\phi_{(1,2,0)}(G)
\ne
\phi_{(1,2,0)}(G').
\]

This example establishes that shortest-path connectors
can increase structural discrimination beyond bounded-radius
NSPDK. Whereas pairwise distances describe the separation
between nodes, connector signatures encode additional
information about the structure connecting them.The result demonstrates additional discriminative capacity
for the specified structural parameters, rather than
completeness as a graph isomorphism test or general
dominance over other graph representation methods.
\section{Diagonal Dominance and Robustness Analysis}
\label{sec:Diagonal}

A normalized Gram matrix is nearly diagonal when off-diagonal similarities
are near zero. Exact matching can give low feature overlap;
\citet{yanardag2015deep} discuss this issue for graph substructure features.

\begin{figure}[!t]
  \centering
  \begin{adjustbox}{max width=0.97\linewidth}
    \begin{minipage}{\linewidth}
      \Subpanel[height=0.78\linewidth,keepaspectratio]{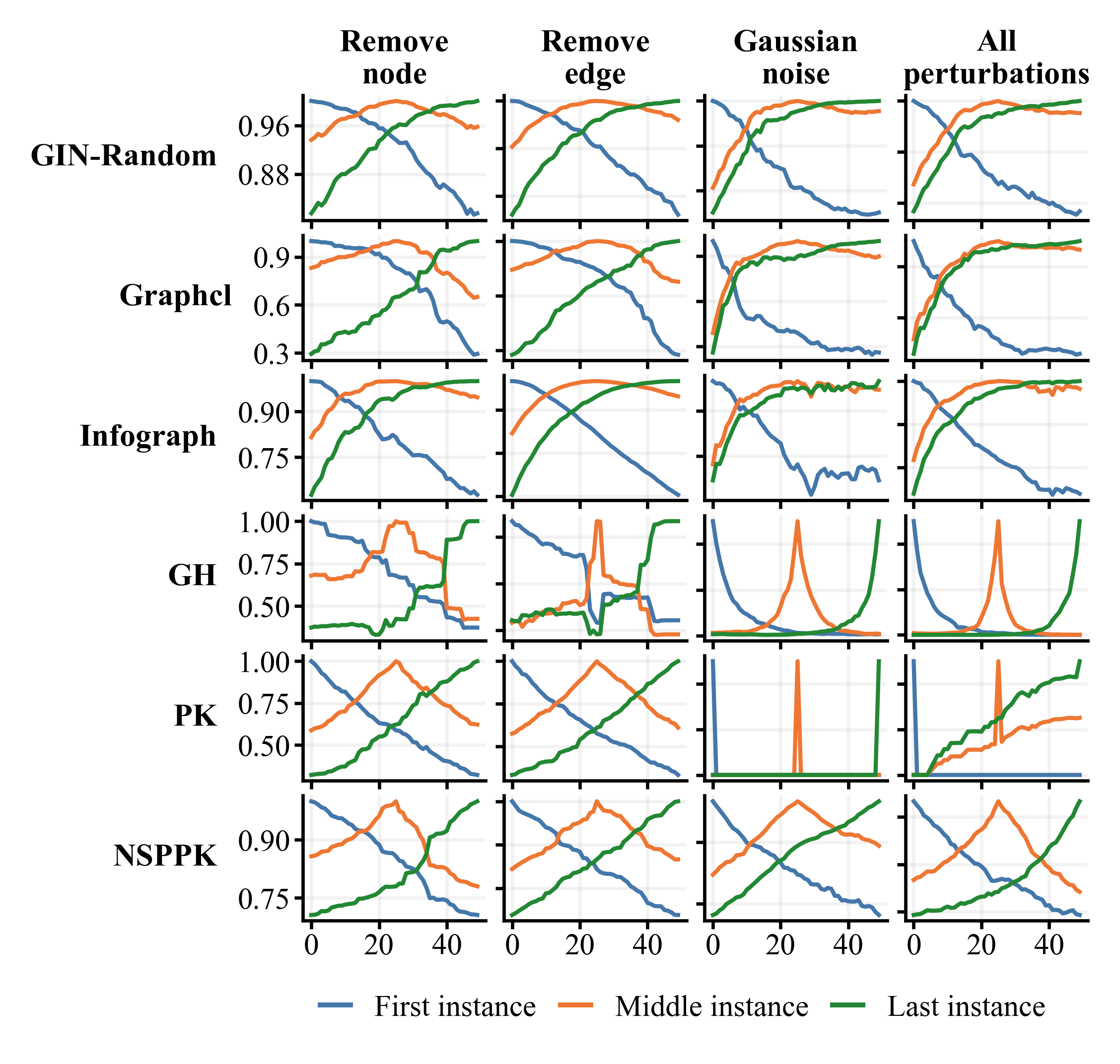}{a}
        {Instance similarity curves}{fig:inst-curves}
      \hfill
      \Subpanel[height=0.78\linewidth,keepaspectratio]{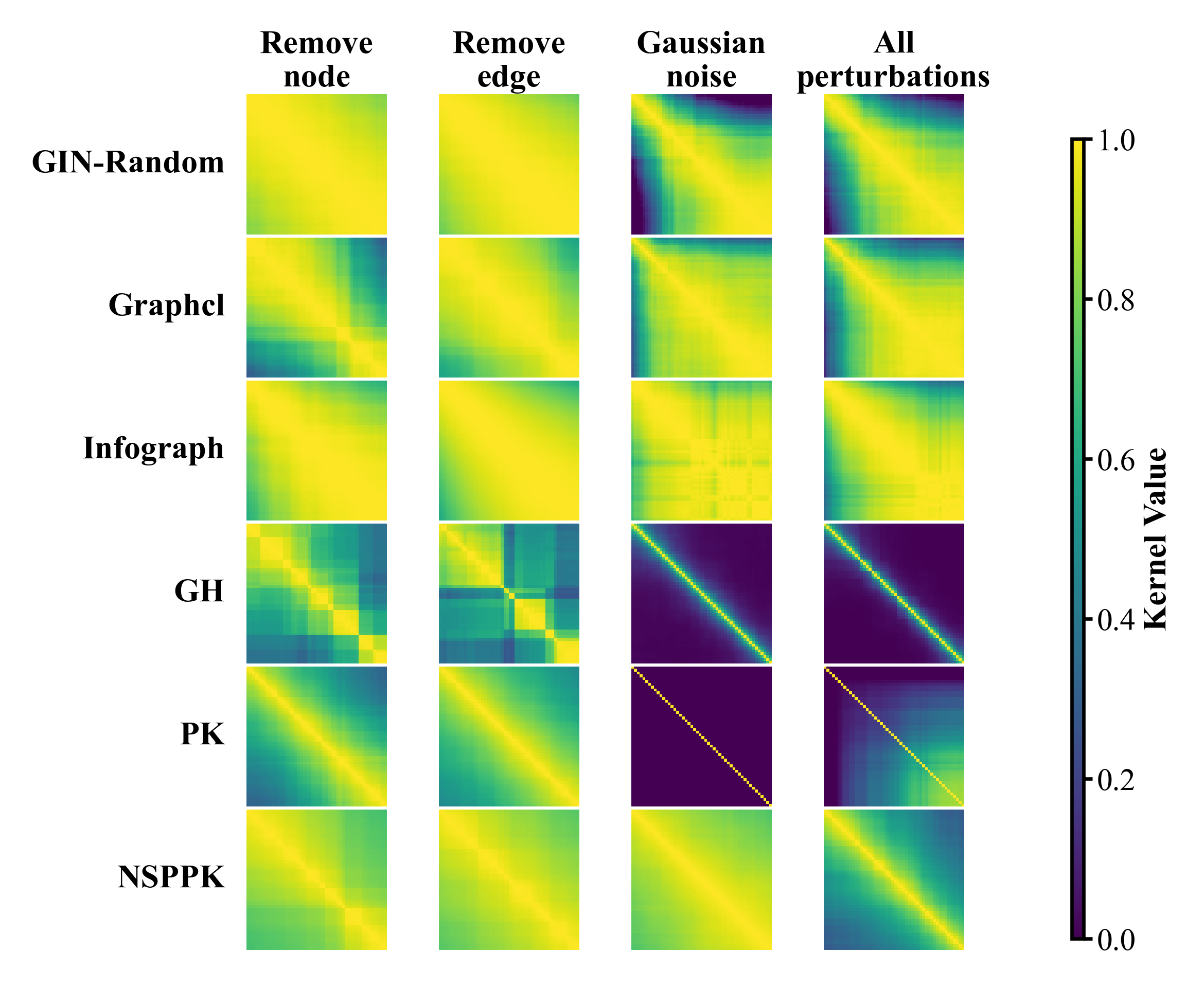}{b}
        {Pairwise similarity matrices}{fig:pairwise}

      \vspace{2mm}

      \Subpanel[height=0.78\linewidth,keepaspectratio,trim=8 6 8 10,clip]{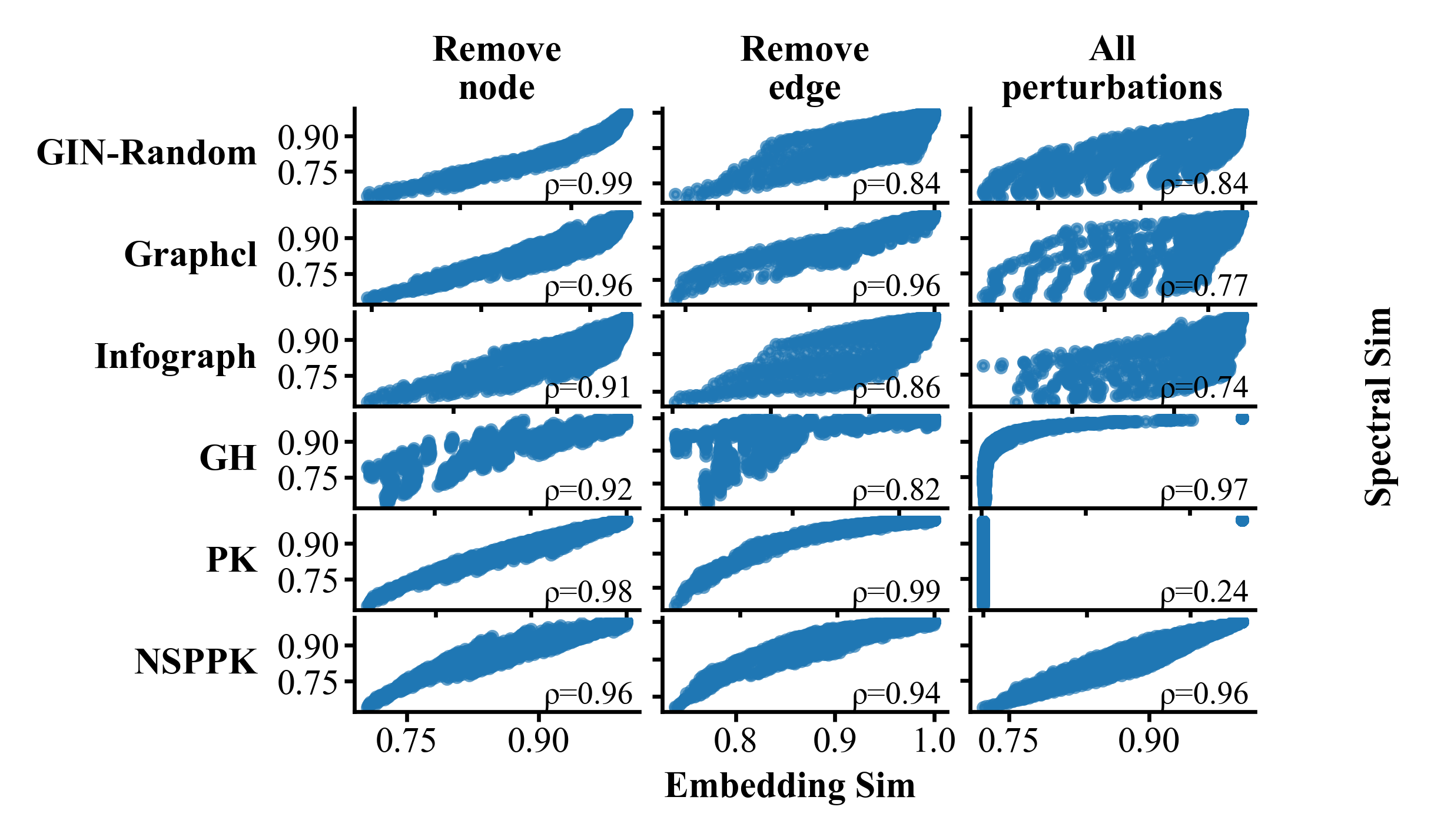}{c}
        {Embedding vs.\ spectral similarity}{fig:emb-vs-spec}
      \hfill
      \Subpanel[height=0.78\linewidth,keepaspectratio]{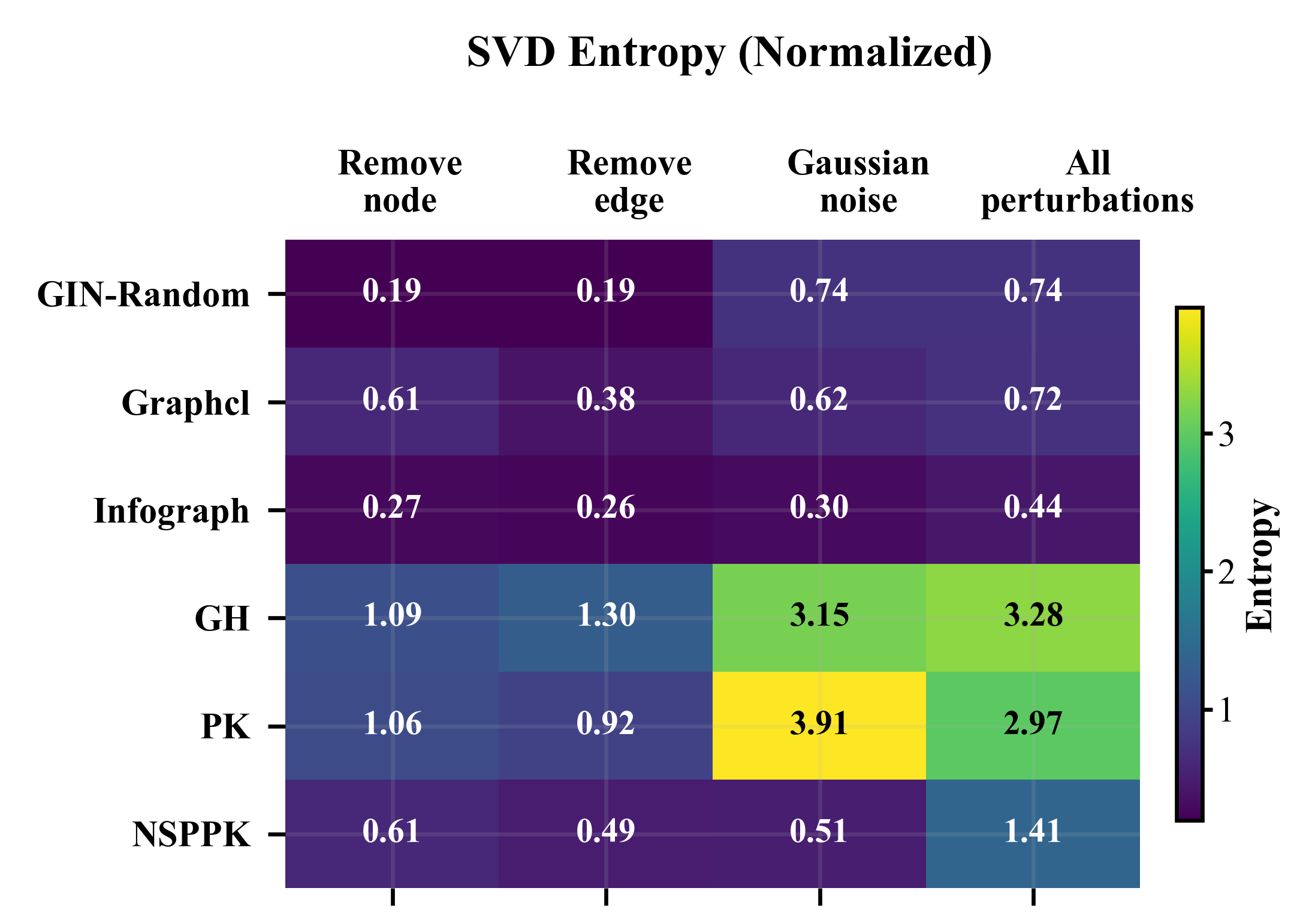}{d}
        {SVD entropy matrices}{fig:svd-entropy}
    \end{minipage}
  \end{adjustbox}

  \caption{\textbf{Unified robustness analysis of graph similarity methods under perturbations.}
  (a) Similarity to the first, middle, and last graphs in each perturbation sequence.
  (b) Normalized pairwise similarity matrices, showing off-diagonal coupling and smoothness patterns.
  (c) Correlation between kernel similarity and spectral similarity (Spearman $\rho$).
  (d) Entropy of the normalized singular-value distribution.}
  \label{fig:perturbation_panel}
\end{figure}

We start from a prototype graph with 200 nodes and generate a sequence of
50 perturbations in each of four regimes: node deletion, edge removal,
Gaussian noise on five-dimensional node attributes, and a combination of
the three. The structural perturbations preserve connectivity. We compare
NSPPK with a randomly initialized GIN architecture~\citep{xu2019powerful},
GraphCL~\citep{you2020graph}, InfoGraph~\citep{sun2019infograph},
GraphHopper~\citep{feragen2013scalable}, and the Propagation
Kernel~\citep{neumann2016propagation}. Appendix~\ref{appendix:robustness}
lists the configurations.

To separate feature-vector direction from changes in feature-vector norm, all
similarities in this analysis use the normalized Gram matrix
\[
\widehat K_{ij}=\frac{K_{ij}}{\sqrt{K_{ii}K_{jj}}},
\]
with zero similarity assigned if either diagonal term is zero. Raw feature norms
are not used as a robustness score.

Figure~\ref{fig:perturbation_panel}(a) shows similarities to the first,
middle, and last graphs in each sequence. The first-instance NSPPK curve
generally declines across the four regimes. Under Gaussian noise, the
GraphHopper and Propagation Kernel similarities to the first instance
fall much more sharply.

Figure~\ref{fig:perturbation_panel}(b) shows the normalized pairwise similarities among the perturbed graphs. Under Gaussian noise, GraphHopper and the attribute Propagation Kernel retain high similarity to each graph itself but assign little similarity to other graphs. Their matrices therefore have bright diagonals and faint off-diagonal entries, indicating weak similarity across distinct instances. The neural methods retain stronger similarities between different graphs, producing smoother matrices in these experiments.

Figure~\ref{fig:perturbation_panel}(c) reports Spearman correlations with
Laplacian spectral similarity for node deletion, edge removal, and combined
perturbations. NSPPK obtains $\rho=0.96$, $0.94$, and $0.96$, respectively.
For the combined regime the Propagation Kernel obtains $0.24$ and GraphHopper
obtains $0.97$.

Finally, we report the entropy of the singular-value distribution of
$\widehat K$. If $s_i$ are its singular values and
$p_i=s_i/\sum_j s_j$, then
$\Xi_{\mathrm{svd}}=-\sum_{i:p_i>0}p_i\log p_i$ and
$\exp(\Xi_{\mathrm{svd}})$ is the corresponding effective rank
(Figure~\ref{fig:perturbation_panel}(d)). For a nonzero $N\times N$ matrix,
$0\leq\Xi_{\mathrm{svd}}\leq\log N$. Entropy is zero for a rank-one matrix
and $\log N$ for the identity. Under Gaussian noise, the Propagation Kernel has high
entropy ($3.91$) and a nearly diagonal heatmap. In this case, a flat spectrum
accompanies near-zero similarities between distinct graphs.

Across the tested sequences, NSPPK generally assigned higher similarity to
graphs closer together in a sequence. Under Gaussian attribute noise, it
retained substantial similarity between distinct graphs. Under combined
perturbations, its ordering of graph pairs by similarity closely followed
the ordering given by Laplacian spectral similarity.

\FloatBarrier
\section{Experiments}
\label{sec:experiments}

\subsection{Small-Scale Datasets}
\label{sec:small_datasets}

We evaluate six node-attributed graph classification benchmarks from
TUDataset~\citep{morris2020tudataset}. ENZYMES and PROTEINS\_full are protein
datasets~\citep{borgwardt2005protein}. BZR, COX2, and DHFR are molecular
activity datasets~\citep{sutherland2003spline}, and SYNTHETICnew is a synthetic
benchmark used by \citet{morris2016fasterkernelsgraphscontinuous}.

\begin{table}[!hb]
\centering
\caption{Classification accuracy (\%) with node attributes (+) and average rank.}
\label{tab:with_attr}
\scriptsize
\setlength{\tabcolsep}{3.2pt}
\begin{adjustbox}{max width=\textwidth}
\begin{tabular}{lccccccc}
\toprule
\textbf{Method} & \textbf{SYNTHETICnew} & \textbf{DHFR} & \textbf{BZR} & \textbf{COX2} & \textbf{ENZYMES} & \textbf{PROTEINS\_full} & \textbf{Avg Rank} \\
\midrule
SM                      & \TO & \TO & \mps{83.96}{3.85} & \mps{78.81}{4.49} & \TO & \TO & 15.50 \\
SP                      & \TO & \TO & \TO & \TO & \TO & \TO & 18.67 \\
ML                      & \mps{50.33}{9.00} & \mps{78.57}{4.21} & \mps{82.96}{5.24} & \mps{77.53}{5.53} & \mps{36.33}{4.33} & \mps{72.06}{3.60} & 11.75 \\
PK                      & \mps{54.33}{10.11} & \mps{60.98}{0.48} & \mps{78.77}{1.01} & \mps{78.16}{8.07} & \mps{20.67}{2.70} & \mps{59.70}{0.16} & 14.42 \\
HSPPK\_WL               & \mps{60.33}{6.74} & \mps{77.38}{5.44} & \mps{85.67}{3.46} & \mps{80.30}{4.67} & \mps{60.17}{6.26} & \mps{72.96}{4.84} & 6.58\\
HSPPK\_SP               & \mps{57.00}{7.81} & \mps{77.40}{3.85} & \mps{84.49}{5.04} & \mps{80.31}{5.70} & \mps{58.50}{5.55} & \mps{68.55}{5.24} & 8.33 \\
GH                      & \mps{77.33}{7.42} & \mps{78.19}{5.47} & \mps{85.94}{5.17} & \mps{78.90}{2.95} & \mps{66.50}{6.17} & \mps{72.06}{3.64} & 5.25 \\
NP                      & \textbf{\mps{99.00}{2.13}} & \mps{80.42}{4.57} & \mps{86.18}{5.53} & \mps{78.16}{4.47} & \mps{43.00}{5.76} & \mps{64.62}{5.13} & 7.42 \\
linearFGW-RAW           & \mps{62.00}{8.46} & \mps{60.85}{3.75} & \mps{76.34}{5.68} & \mps{77.93}{3.68} & \mps{56.67}{7.07} & \mps{69.47}{0.91} & 12.17 \\
linearFGW-WL1           & \mps{72.33}{8.70} & \mps{59.94}{4.11} & \mps{78.52}{3.99} & \mps{72.13}{7.40} & \mps{47.00}{4.70} & \mps{59.66}{0.38} & 14.50 \\
linearFGW-WL2           & \mps{71.33}{7.48} & \mps{57.82}{5.48} & \mps{78.51}{2.98} & \mps{75.16}{3.34} & \mps{41.00}{7.57} & \mps{59.75}{0.43} & 14.67 \\
WL (disc.)              & \mps{88.33}{5.63} & \mps{77.26}{5.81} & \mps{83.71}{4.83} & \mps{77.10}{5.59} & \mps{50.67}{7.03} & \mps{71.16}{4.03} & 10.08 \\
WLOA                    & \mps{85.33}{5.62} & \mps{75.00}{3.23} & \mps{84.20}{4.47} & \mps{74.54}{5.40} & \mps{66.33}{5.21} & \mps{71.16}{1.97} & 9.08 \\
WWL                     & \mps{58.33}{7.78} & \mps{77.12}{3.58} & \mps{86.45}{4.50} & \mps{79.03}{3.84} & \textbf{\mps{73.67}{5.26}} & \textbf{\mps{77.18}{5.27}} & 5.33 \\
SWWL                    & \mps{55.33}{12.31} & \mps{77.64}{5.51} & \mps{84.92}{3.10} & \textbf{\mps{81.39}{5.24}} & \mps{64.83}{6.03} & \mps{74.21}{3.15} & 5.92 \\
MMD-GK                  & \mps{55.33}{6.00} & \mps{78.58}{4.10} & \mps{84.19}{5.01} & \mps{77.32}{4.99} & \mps{49.33}{4.42} & \mps{70.98}{3.95} & 10.58 \\
NSPDK (disc.)           & \mps{96.33}{3.14} & \mps{82.93}{5.56} & \mps{85.70}{3.90} & \mps{80.30}{4.15} & \mps{52.67}{4.67} & \mps{72.87}{1.51} & 4.92 \\
\rowcolor{gray!10}
NSPPK (ours)            & \textbf{\mps{99.00}{1.52}} & \textbf{\mps{82.95}{4.57}} & \textbf{\mps{87.17}{3.58}} & \mps{81.16}{2.30} & \mps{60.50}{5.38} & \mps{74.66}{3.81} & \textbf{2.08} \\
\midrule
\emph{Attributes only}  & \mps{54.33}{9.55} & \mps{60.98}{4.80} & \mps{78.77}{1.01} & \mps{78.16}{8.07} & \mps{55.67}{5.38} & \mps{62.80}{2.52} & 12.75 \\
\bottomrule
\end{tabular}
\end{adjustbox}

\vspace{0.5ex}
{\footnotesize\emph{Note:} TIMEOUT denotes a run exceeding 24 hours for one fold and is assigned the tied worst rank on that dataset. Avg Rank is computed across all six datasets (lower is better); ``Attributes only'' participates like any other method.}
\end{table}

\begin{table}[!tb]
\centering
\caption{Classification accuracy (\%) with node attributes in the neural-baseline
comparison. Ranks use the displayed mean accuracies, with ties assigned the
average rank.}
\label{tab:nn_with_attr}
\scriptsize
\setlength{\tabcolsep}{3.2pt}
\begin{adjustbox}{max width=\textwidth}
\begin{tabular}{lccccccc}
\toprule
\textbf{Method} & \textbf{SYNTHETICnew} & \textbf{DHFR} & \textbf{BZR} & \textbf{COX2} & \textbf{ENZYMES} & \textbf{PROTEINS\_full} & \textbf{Avg Rank} \\
\midrule
DGCNN                & \mps{46.67}{5.63} & \mps{68.77}{5.52} & \mps{79.40}{3.32} & \mps{77.15}{0.06} & \mps{33.33}{9.37} & \mps{73.86}{3.56} & 8.83 \\
GraphSAGE            & \mps{76.67}{7.70} & \mps{76.46}{4.13} & \mps{83.70}{4.44} & \mps{80.93}{0.07} & \mps{65.00}{5.73} & \mps{75.02}{3.29} & 4.42 \\
InfoGraph            & \mps{65.00}{16.05} & \mps{73.01}{6.03} & \mps{79.01}{3.42} & \mps{77.77}{14.20} & \mps{53.33}{7.84} & \mps{66.37}{6.25} & 8.08 \\
GIN                  & \mps{83.67}{5.92} & \mps{76.85}{7.23} & \mps{84.17}{6.14} & \mps{81.80}{6.14} & \textbf{\mps{68.30}{5.43}} & \mps{62.10}{5.26} & 4.25\\
GraphCL              & \mps{67.00}{9.48} & \mps{80.34}{6.95} & \mps{84.17}{3.62} & \mps{80.34}{6.95} & \mps{48.17}{6.93} & \mps{75.82}{2.73} & 4.25 \\
GNN                  & \mps{64.67}{7.92} & \mps{76.59}{6.97} & \mps{85.66}{4.60} & \mps{79.92}{7.08} & \mps{65.17}{8.41} & \mps{66.57}{6.10} & 5.00 \\
FNP                  & \mps{53.33}{11.16} & \mps{70.76}{3.99} & \mps{79.49}{3.94} & \mps{78.22}{7.01} & \mps{32.17}{12.98} & \mps{70.17}{3.15} & 8.25 \\
PNA                  & \mps{55.67}{20.66} & \mps{61.50}{5.42} & \mps{79.01}{4.33} & \mps{78.22}{7.02} & \mps{20.83}{7.12} & \mps{75.11}{3.60} & 8.17 \\
PDF                  & \mps{97.67}{2.60} & \mps{77.37}{4.50} & \mps{83.68}{3.81} & \mps{82.23}{7.00} & \mps{65.00}{4.65} & \mps{72.14}{4.48} & 3.75 \\
\rowcolor{gray!10}
NSPPK feat.\ (XGBoost) & \textbf{\mps{98.67}{2.21}} & \textbf{\mps{83.74}{4.14}} & \textbf{\mps{88.66}{2.89}} & \textbf{\mps{82.90}{4.39}} & \mps{60.17}{6.30} & \textbf{\mps{77.37}{4.80}} & \textbf{1.67} \\
\midrule
\emph{Attributes only} & \mps{54.33}{9.55} & \mps{60.98}{ 4.80}& \mps{78.77}{1.01} & \mps{78.16}{8.07} & \mps{55.67}{5.38} & \mps{62.80}{2.52} & 9.33 \\
\bottomrule
\end{tabular}
\end{adjustbox}
\end{table}
\textbf{Kernel baselines.}
We compare Subgraph Matching (SM)~\citep{kriege2012subgraph},
Shortest Path (SP)~\citep{borgwardt2005shortest},
Multiscale Laplacian (ML)~\citep{kondor2016multiscalelaplaciangraphkernel},
Propagation Kernel (PK)~\citep{neumann2016propagation},
GraphHopper (GH)~\citep{feragen2013scalable}, and
Neighborhood Preserving Kernel (NP)~\citep{salim2022neighborhood}.
HSPPK-SP and HSPPK-WL denote hash graph kernels with Shortest Path and
Weisfeiler--Lehman Subtree base kernels, respectively~\citep{morris2016fasterkernelsgraphscontinuous}.
We also include the Weisfeiler--Lehman Subtree kernel (WL)~\citep{shervashidze2011weisfeiler}
with discretized inputs, Weisfeiler--Lehman Optimal Assignment (WLOA)~\citep{kriege2017validoptimalassignmentkernels},
Wasserstein Weisfeiler--Lehman (WWL)~\citep{togninalli2019wwl},
Sliced Wasserstein Weisfeiler--Lehman (SWWL)~\citep{carpintero2024swwl},
Maximum Mean Discrepancy Graph Kernel (MMD-GK)~\citep{mmdgk_iclr2024}, and
the Neighborhood Subgraph Pairwise Distance Kernel (NSPDK)~\citep{costa2010fast},
including a discretized variant.
For Linear Fused Gromov--Wasserstein (linearFGW)~\citep{nguyen2023linear},
RAW uses the input node features. WL1 and WL2 use node features after one and
two rounds of Weisfeiler--Lehman neighborhood aggregation, respectively.
All kernels use support vector machine classifiers (LIBSVM~\citep{chang2011libsvm}).

\textbf{Neural baselines}: DGCNN~\citep{zhang2018endtoend}, GraphSAGE~\citep{hamilton2017graphsage}, InfoGraph~\citep{sun2019infograph}, GIN~\citep{xu2019powerful}, GraphCL~\citep{you2020graph}, Attentive FP (labeled FNP in the experiments)~\citep{xiong2019attentivefp}, PNA~\citep{corso2020pna}, and PDF~\citep{yang2023bettergraphrepresentationlearning}.  

\textbf{Protocol.}
Following the separation of model selection and assessment in
\citet{errica2020fair}, we use 10 outer folds and set aside 10\% of each
training fold for validation. We report three
comparison tracks separately. In the \emph{kernel comparison}, each method
produces a Gram matrix and is evaluated with LIBSVM; the SVM parameter $C$ and
baseline-specific hyperparameters are selected on the validation split. In the
\emph{explicit-feature comparison}, NSPPK produces sparse feature vectors that
are passed to standard downstream classifiers, primarily XGBoost, with the
classifier settings selected on validation data. In the \emph{neural
comparison}, GNN baselines are trained with early stopping and validation-based
model selection. We also test (i) \emph{attributes only}, where graphs are
represented by summed node attributes, and (ii) \emph{structure only}, where
attributes are removed (Appendix~\ref{appendix:no_attr}).

\textbf{NSPPK configuration.}
Unless stated otherwise, NSPPK uses the same structural configuration across all
small datasets ($R{=}1$, $D{=}4$, $c{=}1$, hash bits $b{=}16$, degree cutoff
$\tau{=}8$ when thresholding is enabled). Thus NSPPK is not tuned separately per
dataset in the main small-scale experiments, although the method does expose
these structural and implementation hyperparameters.

\textbf{Runtime.}  
The small-dataset runs used a Slurm-managed cluster with Intel Xeon Gold
5317 processors and 64 GB of RAM. Each processor has 12 physical cores and
24 hardware threads. Kernel runs used one CPU core, and GNN runs used CPU
multithreading. We report a timeout when a kernel fold exceeds 24 hours.
Runtime comparisons use the settings of each track.

\begin{table}[htbp]
\centering
\caption{Kernel runtimes (seconds) with node attributes (\textbf{lower is better}).}
\label{tab:with_attr_runtime_kernels}
\scriptsize
\setlength{\tabcolsep}{3.5pt}
\begin{adjustbox}{max width=\textwidth}
\begin{tabular}{lcccccccc}
\toprule
\textbf{Method} & \textbf{SYNTHETICnew} & \textbf{DHFR} & \textbf{BZR} & \textbf{COX2} & \textbf{ENZYMES} & \textbf{PROTEINS\_full} & \textbf{Average} & \textbf{Rank}\\
\midrule
SM & \TO & \TO & 12274.00\,s & 25927.96\,s & \TO & \TO & 19100.98\,s & 17.00\\
ML & 1777.93\,s & 1361.75\,s & 849.20\,s & 1155.49\,s & 1295.99\,s & 3628.70\,s & 1678.18\,s & 15.83\\
PK & 1.81\,s & 2.82\,s & 0.79\,s & 1.13\,s & 3.14\,s & 4.48\,s & 2.36\,s & \textbf{1.33}\\
HSPPK\_WL & 49.54\,s & 51.49\,s & 8.63\,s & 16.38\,s & 25.20\,s & 89.90\,s & 40.19\,s & 10.50\\
HSPPK\_SP & 249.75\,s & 63.10\,s & 16.52\,s & 29.22\,s & 28.35\,s & 119.32\,s & 84.38\,s & 12.17\\
GH & 242.05\,s & 318.80\,s & 112.43\,s & 132.63\,s & 365.25\,s & 3647.21\,s & 803.06\,s & 15.00\\
NP & 41.57\,s & 217.54\,s & 42.32\,s & 69.80\,s & 17.69\,s & 49.58\,s & 73.08\,s & 11.67\\
linearFGW-RAW & 6.82\,s & 22.14\,s & 6.63\,s & 9.84\,s & 8.73\,s & 69.67\,s & 20.64\,s & 6.50\\
linearFGW-WL1 & 6.94\,s & 22.58\,s & 5.89\,s & 10.75\,s & 8.29\,s & 67.19\,s & 20.27\,s & 6.50\\
linearFGW-WL2 & 5.95\,s & 25.23\,s & 7.02\,s & 10.92\,s & 10.26\,s & 71.63\,s & 21.84\,s & 7.50\\
WL (disc.) & 12.33\,s & 0.21\,s & 4.00\,s & 14.00\,s & 9.00\,s & 20.00\,s & 9.92\,s & 5.83\\
WLOA & 0.99\,s & 10.49\,s & 0.83\,s & 3.27\,s & 4.24\,s & 12.67\,s & 5.42\,s & 3.08\\
WWL & 29.96\,s & 105.03\,s & 14.71\,s & 40.89\,s & 32.54\,s & 134.95\,s & 59.68\,s & 11.67\\
SWWL & 3.52\,s & 6.71\,s & 3.16\,s & 3.84\,s & 5.03\,s & 10.92\,s & 5.53\,s & 3.50\\
NSPDK (disc.) & 6.50\,s & 10.49\,s & 4.69\,s & 2.64\,s & 3.18\,s & 6.13\,s & 5.61\,s & 3.58\\
MMD-GK & 38.07\,s & 156.26\,s & 40.48\,s & 57.43\,s & 86.43\,s & 305.15\,s & 113.97\,s & 13.00\\
\rowcolor{gray!10}
\textbf{NSPPK (ours)} & 34.81\,s & 36.19\,s & 12.75\,s & 12.75\,s & 26.45\,s & 5.73\,s & 21.45\,s & 8.33\\
\bottomrule
\end{tabular}
\end{adjustbox}
\end{table}

\begin{table}[htbp]
\centering
\caption{Runtimes (seconds) with node attributes in the neural-baseline
comparison. The final columns average the displayed times and their
within-dataset ranks. Faster runs receive lower ranks.}
\label{tab:with_attr_runtime_nns}
\scriptsize
\setlength{\tabcolsep}{3.5pt}
\begin{adjustbox}{max width=\textwidth}
\begin{tabular}{lcccccccc}
\toprule
\textbf{Method} & \textbf{SYNTHETICnew} & \textbf{DHFR} & \textbf{BZR} & \textbf{COX2} & \textbf{ENZYMES} & \textbf{PROTEINS\_full} & \textbf{Average} & \textbf{Rank}\\
\midrule
DGCNN & 443.59\,s & 2956.95\,s & 304.88\,s & 704.74\,s & 948.88\,s & 1512.29\,s & 1145.22\,s & 9.83\\
GraphSAGE & 139.94\,s & 101.76\,s & 175.30\,s & 118.97\,s & 317.84\,s & 394.02\,s & 207.97\,s & 7.33\\
InfoGraph & 39.25\,s & 163.79\,s & 40.40\,s & 111.40\,s & 59.95\,s & 85.72\,s & 83.42\,s & 5.33\\
GIN & 474.24\,s & 1066.25\,s & 302.44\,s & 579.86\,s & 369.06\,s & 742.46\,s & 589.05\,s & 9.17\\
GraphCL & 11.33\,s & 5.11\,s & 4.10\,s & 5.11\,s & 15.51\,s & 20.04\,s & 10.20\,s & \textbf{1.17}\\
GNN & 165.92\,s & 242.21\,s & 100.28\,s & 153.89\,s & 207.18\,s & 371.30\,s & 206.80\,s & 7.50\\
FNP & 49.97\,s & 71.08\,s & 18.72\,s & 22.23\,s & 60.51\,s & 57.51\,s & 46.67\,s & 4.50\\
PNA & 16.67\,s & 27.45\,s & 4.38\,s & 3.86\,s & 17.33\,s & 22.14\,s & 15.30\,s & 1.83\\
PDF & 20.59\,s & 57.84\,s & 21.10\,s & 20.13\,s & 61.43\,s & 76.70\,s & 42.96\,s & 4.17\\
\rowcolor{gray!10}
\textbf{NSPPK feat.\ (XGB)} & 39.35\,s & 33.10\,s & 19.38\,s & 20.62\,s & 39.54\,s & 131.96\,s & 47.32\,s & 4.17\\
\bottomrule
\end{tabular}
\end{adjustbox}
\end{table}

\textbf{Results.}
NSPPK has the best average rank in the attributed kernel comparison
($2.08$). It leads on DHFR and BZR and ties NP on SYNTHETICnew.
NSPPK with XGBoost has the best average rank in the comparison
with neural baselines ($1.67$) and the highest mean accuracy on five
of the six datasets. The attributes-only control scores below NSPPK on all
six datasets. Appendix~\ref{appendix:cd} reports the statistical comparisons
of accuracy ranks.

\textbf{Runtime.}
Tables~\ref{tab:with_attr_runtime_kernels} and~\ref{tab:with_attr_runtime_nns}
report runtimes with attributes. In the single-core kernel comparison, NSPPK
averages 21.45 seconds and GraphHopper averages 803.06 seconds. In the
neural comparison, GraphCL and PNA report lower mean runtimes than NSPPK
with XGBoost. Appendix~\ref{appendix:tradeoff} plots accuracy against runtime
for each dataset and across all six datasets.

\textbf{Parallel scalability.}
NSPPK feature extraction operates independently over graphs. On the
preprocessed QM9 collection, vectorization drops from over one hour on one
core to under ten minutes on 112 logical cores
(Figure~\ref{fig:appendix_scaling_plot} in Appendix~\ref{appendix:scaling}).

\subsection{Larger-Scale Dataset Experiment}
\label{sec:large_datasets}
\begin{figure}[H]
  \centering
  \includegraphics[width=0.70\textwidth]{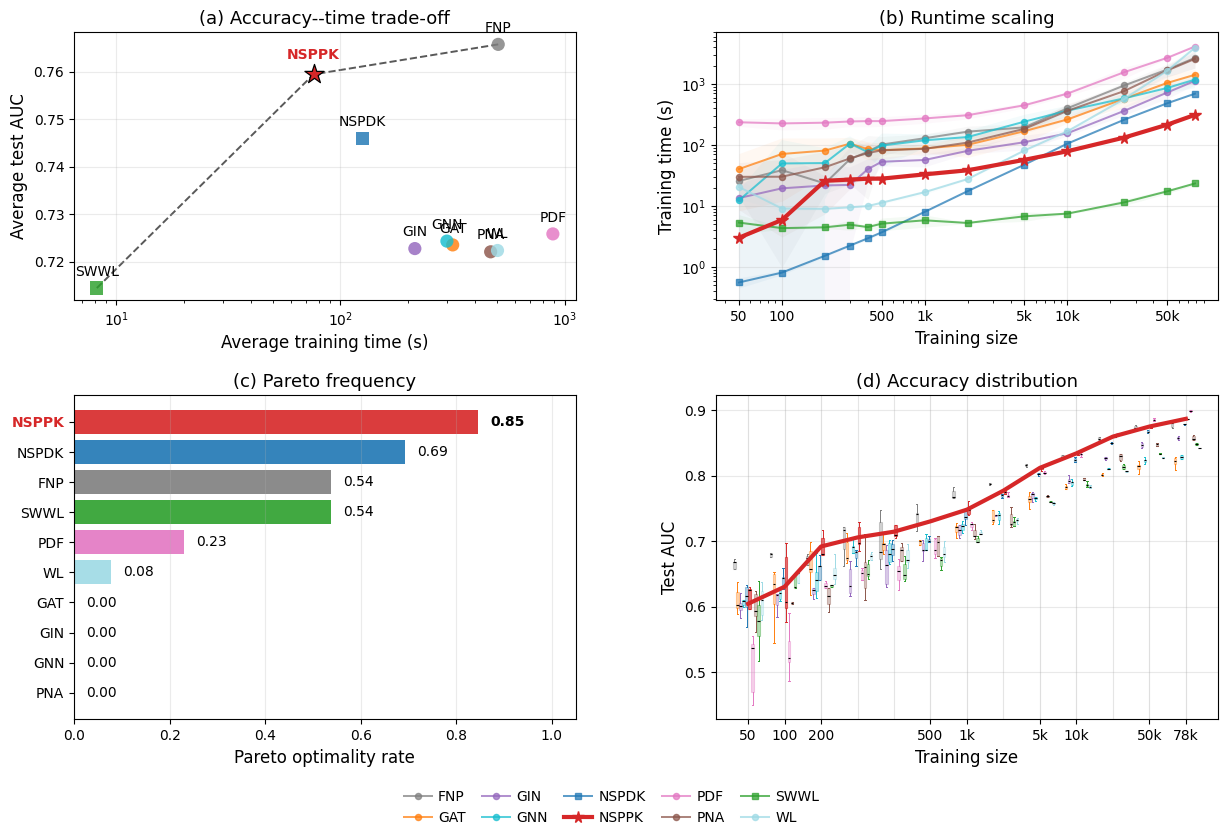}
  \vspace{-4pt}
  \caption{Task~95 results on \texttt{ogbg-molpcba}. (a) Mean test ROC--AUC
  versus mean training time. (b) Training time by training size. (c) Pareto
  frequency based on ROC--AUC and training time. (d) Test ROC--AUC by
  training size.}
  \vspace{-8pt}
  \label{fig:large_experiment}
\end{figure}

We evaluate the 95th \texttt{ogbg-molpcba} task, which predicts activity in a cell-based screen for human TDP1 inhibitors conducted in the presence of camptothecin (PubChem AID 686979) using balanced subsets of
approximately 78k graphs. Appendix~\ref{appendix:task95} gives
the full protocol and runtime decomposition. Against neural and scalable kernel
baselines, NSPPK shows strong sample efficiency and reaches the Pareto frontier
for most configurations (Figure~\ref{fig:large_experiment}).
Table~\ref{tab:task95_efficiency} reports end-to-end time separately and
includes both feature extraction and training.

\section{Ablation Analysis}
\label{sec:ablation_sensitivity_density}
\label{sec:component_ablation}

We ablate distance features, the union-of-shortest-paths connector, and
high-degree thresholding under a fixed configuration
($R=1,D=4,c=1,b=16,\tau=8$) with a random forest classifier. The full model is
best or tied-best on most benchmarks (Table~\ref{tab:ablation}); removing
distance or connector features generally reduces accuracy, whereas thresholding
has a dataset-dependent effect. Full protocol details are given in
Appendix~\ref{appendix:ablation-details}.

\begin{table}[H]
\centering
\caption{Component ablation on six benchmarks. Results are mean test accuracy
(\%) $\pm$ standard deviation over 10 folds; settings and classifier details are
given in Appendix~\ref{appendix:ablation-details}.}
\label{tab:ablation}
\scriptsize
\setlength{\tabcolsep}{3pt}
\begin{adjustbox}{max width=\textwidth}
\begin{tabular}{lcccccc}
\toprule
\textbf{Setting} & \textbf{ENZYMES} & \textbf{BZR} & \textbf{PROTEINS\_full} & \textbf{DHFR} & \textbf{SYNTHETICnew} & \textbf{COX2} \\
\midrule
No Distance        & 57.17 $\pm$ 6.99 & 85.69 $\pm$ 5.52 & \textbf{76.37 $\pm$ 3.85} & 80.17 $\pm$ 4.35 & 98.33 $\pm$ 2.24 & 81.16 $\pm$ 2.45 \\
No Path            & 52.83 $\pm$ 6.28 & 86.98 $\pm$ 3.72 & 76.10 $\pm$ 4.11 & 76.85 $\pm$  2.58 & 98.67 $\pm$ 1.63 & 80.73 $\pm$ 2.30 \\
No Threshold       & 55.50 $\pm$ 4.48 & 87.42 $\pm$ 2.25 & 75.92 $\pm$ 5.05 & \textbf{83.34 $\pm$ 3.10} & \textbf{99.00 $\pm$ 1.53} & 81.81 $\pm$ 3.78 \\
\textbf{Full NSPPK}& \textbf{57.83 $\pm$ 5.43} & \textbf{87.90 $\pm$ 3.56} & 76.20 $\pm$ 3.74 & 82.95 $\pm$ 4.57 & \textbf{99.00 $\pm$ 1.53} & \textbf{82.46 $\pm$ 3.74} \\
\bottomrule
\end{tabular}
\end{adjustbox}
\end{table}

Appendix~\ref{appendix:sensitivity-results} further shows comparable accuracy
across several classifiers, only $1$--$2$ percentage-point variation across the
COX2 structural-parameter sweeps, and the measured runtime increase on dense
graphs. These results reduce, but do not eliminate, sensitivity to classifier
choice and structural tuning within the evaluated settings.
To examine the computational behavior of NSPPK beyond
sparse graph structures, we additionally measure feature
extraction time on Erdős--Rényi graphs with a fixed number
of nodes and increasing expected average degree.
This experiment investigates the effect of graph density
on vectorization cost and complements the theoretical
complexity analysis. Experimental settings and results
are provided in Appendix~\ref{appendix:sensitivity-results}.

\section{Conclusion}
\label{sec:Conclusion}
In this work, we introduced the Neighborhood Subgraph Pairwise
Path Kernel (NSPPK), a deterministic graph kernel that maps graphs
with continuous node attributes into high-dimensional sparse
feature vectors. Its core structural behavior is controlled by just
three integer parameters. The method combines structural feature
hashing with the direct aggregation of continuous node attributes,
producing explicit representations that can be computed once per
graph and reused across learning tasks. In our experiments, a
single structural configuration performed competitively across all
six graph classification benchmarks, without dataset-specific
structural parameter tuning.

NSPPK also supports efficient, parallel graph vectorization.
Feature extraction is independent across graphs, making it
straightforward to distribute across processors and well suited to
rapid experiments on large collections of sparse graphs. In the
QM9 experiment, vectorization time fell from over an hour on one
core to under ten minutes on 112 logical cores. Theoretical
analysis establishes linear scaling in the number of nodes when
structural parameters are fixed and local adjacency sizes are
bounded. Shortest-path connectors also distinguish graph structures
that bounded-radius NSPDK cannot distinguish, while perturbation
experiments show that NSPPK retains informative similarities under
structural and attribute modifications.

Across six graph classification benchmarks, NSPPK achieves the
best average accuracy rank among the evaluated graph kernels.
Combined with XGBoost, its explicit features achieve the highest
mean accuracy on five of the six datasets in the neural-baseline
comparison. Ablation results support the contribution of distance
and connector features. The MolPCBA case study extends the
evaluation to larger training sets and shows how NSPPK performs as
the amount of training data grows.

\textbf{Limitations and future work.}
The computational cost of NSPPK depends on graph connectivity
and the extent of the extracted structural patterns. Although
the method is particularly well suited to datasets comprising
small, sparse graphs, increasing graph density or local
connectivity can substantially increase the number of eligible
node pairs and the cost of constructing structural features.
As demonstrated by the density sensitivity experiment, this
can lead to considerable increases in vectorization time.

Overall, NSPPK demonstrates that deterministic graph
vectorization can achieve competitive predictive performance
without requiring trainable graph encoders. Its explicit
feature representation supports both kernel-based learning
and conventional machine learning algorithms, providing a
practical alternative to learned graph embeddings,
particularly for collections of sparse attributed graphs.
\paragraph{Code availability.} The implementation corresponding to the experiments reported in this paper is available in the \href{https://github.com/fabriziocosta/NSPPK/releases/tag/v1.3.6}{NSPPK v1.3.6} release (\url{https://github.com/fabriziocosta/NSPPK/releases/tag/v1.3.6}). Later releases include additional functionality beyond the version used for this paper.

\paragraph{Acknowledgments.}
For the purpose of open access, the authors have applied a Creative
Commons Attribution (CC BY) license to any Author Accepted Manuscript
version arising from this submission.
\bibliographystyle{tmlr}
\bibliography{bib}

\appendix
\section*{Appendix}

\section{Definitions}
\label{sec:definitions}

A \emph{graph} here is a finite, simple, undirected, unweighted graph
$G=(V,E)$, where $E\subseteq\{\{u,v\}\subseteq V:u\ne v\}$.
A \emph{labeled graph} is a graph $G = (V, E)$ equipped with a labeling function
$\ell : V \cup E \to \Sigma$ that assigns each vertex and edge a label from a
discrete alphabet $\Sigma$.
A \emph{node-attributed graph} is a graph $G=(V,E)$ equipped
with an attribute function $a:V\to\mathbb{R}^p$, assigning
each vertex a $p$-dimensional real-valued attribute vector.
Equivalently, the node attributes can be represented by
a matrix $A\in\mathbb{R}^{|V|\times p}$, whose rows
correspond to vertices, with $A_{v,:}=a(v)^\top$.
We denote the resulting attributed graph by $(G,A)$.

For a vertex $v \in V$, the \emph{degree} of $v$ is the number of edges incident
to it,
\[
\deg(v) = \left| \{ u \in V \mid \{v,u\}\in E \} \right|,
\]
and its \emph{(immediate) neighborhood} is
$N(v) = \{ u \in V \mid \{v,u\}\in E \}$. 
A degree cutoff parameter
$\tau$ may be used as an implementation heuristic to suppress expanded rooted
hashes at high-degree centers; unless traversal pruning is explicitly applied,
this does not by itself change the worst-case BFS cost.

A \emph{path} in $G$ is a sequence of distinct vertices $(v_1, v_2, \ldots, v_k)$ such that $\{v_i,v_{i+1}\}\in E$ for all $1 \leq i < k$.
The \emph{length} of the path is the number of edges in the sequence, i.e., $k-1$.

A \emph{shortest path} from $v$ to $u$ is a path with the smallest possible
length among all paths connecting $v$ and $u$.
The \emph{distance} between $v$ and $u$, denoted $d(v, u)$, is the length of a shortest path between them; if no path exists, $d(v, u)$ is defined to be infinite.

The \emph{union of shortest paths} between vertices $v$ and $u$, denoted $U(v, u)$, is the subgraph consisting of all vertices and edges that belong to at least one shortest path from $v$ to $u$ (i.e., the union over all equally-short paths).

The \emph{$r$-hop neighborhood} $N_r(v)$ is the subgraph induced by
$\{u\in V:d(v,u)\le r\}$, rooted at $v$. For a set $S\subseteq V$, $N_r(S)$
is induced by the vertices at distance at most $r$ from $S$.

\emph{Anchors and connector subgraph.}
For an unordered pair of distinct anchors with $d(u,v)\le D$, define
$C_0(u,v)=U(u,v)$ as the path union, and
$C_{r'}(u,v)=N_{r'}(V(U(u,v)))$ for $r'>0$. Disconnected pairs are excluded.

\paragraph{Notation summary.} Unless otherwise specified, we denote by $|V|$ and $|E|$
the numbers of vertices and edges, respectively; $K_{\max} = \max_{v\in V} \deg(v)$ is
the maximum degree and $\tau$ an optional degree cutoff. Distances are
$d(u,v)$, $N_r(v)$ is the $r$-hop neighborhood of $v$, $U(u,v)$ the union of
all shortest $u\leftrightarrow v$ paths, and $C_{r'}(u,v)$ the
connector of radius $r'$. Lowercase $r, d, r'$ denote per-feature radii and
distances, while uppercase $R, D, R'$ are their maximal values. The resulting
feature vector for a graph $G$ under parameters $\theta = (R,D,R')$ is
$\phi_\theta(G)$.

\section{Computational and Space Complexity Analysis}
\label{sec:appendix:complexity}
Let $n=|V|$, $m=|E|$, $H=\max(R,R')$, and $B=2^b$. Use
$\mathcal B_\ell(v)$ and $\mathcal E_\ell(v)$ as in
Section~\ref{sec:complexity}. Then
$P_D(G)=\tfrac12\sum_{v\in V}(|\mathcal B_D(v)|-1)$ is the number of
unordered pairs of distinct vertices within distance $D$.

\subsection{Structural feature extraction}
The rooted-hash and anchor-enumeration traversals take, respectively,
\[
T_H=O\!\left(\sum_v (|\mathcal B_H(v)|+|\mathcal E_H(v)|)\right),
\qquad
T_D=O\!\left(\sum_v (|\mathcal B_D(v)|+|\mathcal E_D(v)|)\right).
\]
These terms count vertex visits and adjacency scans. They exclude sorting
neighbor multisets and BFS shells and assembling their hash inputs. Write
$T_{\mathrm{sort}}$ for that additional rooted-hash work, including base node
hashes. A degree cutoff that suppresses expanded hashes without pruning
traversal can reduce hashing work while leaving the traversal bound intact.

With $L$ connector choices, including the omitted connector,
each node pair contributes at most $(R+1)^2L$ structural
configurations. At fixed hash width, combining the cached
descriptors and updating the feature counts requires
$O((R+1)^2LP_D(G))$ operations. Let $T_{\mathrm{conn}}$
denote the computational cost of constructing the connectors
and computing their structural hashes.
Because $r'=0$ uses the path union,
its cost is included whenever that connector is enabled.
The combined bound is
\[
O\bigl(T_H+T_D+T_{\mathrm{sort}}+(R+1)^2 L P_D(G)+T_{\mathrm{conn}}\bigr).
\]
For fixed structural parameters, fixed hash width, and
bounded maximum node degree, the number and size of
the extracted local structural patterns are bounded
independently of the graph size. Consequently, structural
feature extraction has $O(n)$ time complexity, provided
connector construction is restricted to the corresponding
local neighborhoods.

Without a bound on the maximum degree, $P_D(G)$ can
reach $\Theta(n^2)$, and the computational cost increases
with graph connectivity.

\subsection{Attribute integration and kernel evaluation}
Accumulating $A^\top M_G$ by nonzero entries of $M_G$ takes
$O(p\,\operatorname{nnz}(M_G))$ arithmetic operations, with a further
$O(pB)$ cost if the entire output is initialized densely.
Multiplication with a dense $n\times B$ count matrix takes $O(pnB)$
arithmetic operations using the standard matrix product. With sorted sparse
indices, kernel evaluation is linear in the combined number of nonzeros of
the two compared vectors, structural or attributed.

\subsection{Space complexity}
Cached hashes occupy $O(n(H+1))$ entries. In addition to the input graph, a
serial BFS uses $O(n)$ working entries. If $T_{\mathrm{occ}}$ is the number of emitted
structural occurrences, the structural histogram has at most $\min(B,T_{\mathrm{occ}})$
nonzeros and $M_G$ has at most $\min(nB,2T_{\mathrm{occ}})$ nonzeros. The attributed output
has at most $p\min(B,T_{\mathrm{occ}})$ stored nonzeros, or $pB$ entries if dense.

\section{Datasets}
\label{appendix:datasets}

The six small graph-classification benchmarks are publicly available in
TUDataset~\citep{kersting2016benchmark,morris2020tudataset}. Their nodes carry
discrete labels, continuous attributes, or both.

\textbf{BZR} consists of 405 chemical compounds represented as graphs, where nodes correspond to atoms and edges to chemical bonds. The binary labels derive from measured benzodiazepine receptor activity~\citep{sutherland2003spline}.

\textbf{COX2} contains 467 molecules represented as graphs, labeled according to their activity against cyclooxygenase-2 (COX-2)~\citep{sutherland2003spline}.

\textbf{ENZYMES} consists of 600 protein tertiary structures from the BRENDA database. Each protein belongs to one of six top-level enzyme commission (EC) classes, and the task is to predict the correct class \cite{borgwardt2005protein}.

\textbf{PROTEINS\_full} represents proteins as graphs, where vertices correspond to secondary structure elements. Edges connect vertices that are adjacent in the amino acid sequence or in 3D space. The classification task is to distinguish enzymes from non-enzymes \cite{borgwardt2005protein}.

\textbf{SYNTHETICnew} consists of 300 synthetic graphs in two classes, with
100 vertices per graph and a continuous node attribute. The classes differ in
the amount of topological and attribute perturbation. The original
SYNTHETIC generator is described by \citet{feragen2013scalable};
\citet{morris2016fasterkernelsgraphscontinuous} use the SYNTHETICnew variant.

\textbf{DHFR}
contains 756 molecules represented as graphs, labeled according to their activity against the dihydrofolate reductase (DHFR) enzyme \cite{sutherland2003spline}.

\begin{table}[htbp]
\centering
\caption{Graph classification dataset statistics. The nine MolPCBA node
feature channels are categorical encodings.}
\label{tab:dataset_stats}
\scriptsize
\setlength{\tabcolsep}{4.2pt}
\begin{adjustbox}{max width=\linewidth}
\begin{tabular}{lccccccc}
\toprule
\textbf{Dataset} & \#Graphs & \#Classes / Tasks & Avg.\ |V| & Avg.\ |E| & Avg.\ deg. & Max deg. & Attr.\ dim. \\
\midrule
BZR            & 405     & 2   & 35.75 & 38.36 & 2.15 & 4  & 3  \\
COX2           & 467     & 2   & 41.23 & 43.45 & 2.11 & 4  & 3  \\
ENZYMES        & 600     & 6   & 32.63 & 62.14 & 3.86 & 9  & 18 \\
PROTEINS\_full      & 1113    & 2   & 39.06 & 72.82 & 3.74 & 25 & 29 \\
SYNTHETICnew   & 300     & 2   & 100.00& 196.25& 3.93 & 9  & 1  \\
DHFR           & 756     & 2   & 42.43 & 44.54 & 2.10 & 5  & 3  \\
ogbg-molpcba   & 437{,}929 & 128 & 25.96 & 28.10 & 2.16 & 4  & 9  \\
\bottomrule
\end{tabular}
\end{adjustbox}
\end{table}

\section{Hyperparameters Used for Model Selection in the Graph Classification Task}
\label{appendix:Hyperparameters}

For some kernels, only a subset of the hyperparameters
was optimized, while the rest of the hyperparameters were kept fixed.

\begin{table}[!htbp]
    \centering
    \caption{Neural hyperparameters used for model selection: architecture and search ranges.}
    \label{tab:nn_hyperparameters_architecture}
    \scriptsize
    \renewcommand{\arraystretch}{1.12}
    \setlength{\tabcolsep}{3.5pt}
    \begin{adjustbox}{center,max width=\textwidth}
    \begin{tabular}{lcccccc}
        \toprule
        \textbf{Model} &
        \textbf{Layers} &
        \makecell{\textbf{Convs}\\\textbf{per layer}} &
        \makecell{\textbf{Batch}\\\textbf{size}} &
        \makecell{\textbf{Learning}\\\textbf{rate}} &
        \makecell{\textbf{Hidden}\\\textbf{units}} &
        \textbf{Epochs} \\
        \midrule
        DGCNN & 2,3,4 & 1 & 16 & $1\mathrm{e}{-4}$ & 32, 64 & 1000 \\
        GIN & \makecell{see\\hidden units} & 1 & 32, 128 & $1\mathrm{e}{-2}$ &
        \makecell{32 (5 layers), 64 (5 layers),\\64 (2 layers), 32 (3 layers)} & 1000 \\
        GraphSAGE & 3, 5 & 1 & 16 & $1\mathrm{e}{-2},\,1\mathrm{e}{-3},\,1\mathrm{e}{-4}$ & 32, 64 & 1000 \\
        InfoGraph & 3, 5 & -- & 16, 32 & $1\mathrm{e}{-2},\,1\mathrm{e}{-3}$ & 32, 64, 128 & 100 \\
        GNN & 3, 5 & 1 & 16 & $1\mathrm{e}{-2},\,1\mathrm{e}{-3},\,1\mathrm{e}{-4}$ & 32, 64 & 1000 \\
        GraphCL & 2, 3 & 1 & 32 & $1\mathrm{e}{-3}$ & 64, 128 & 200 \\
        FNP & 2, 3, 4 & 1 & 32 & $1\mathrm{e}{-3}$ & 32, 64 & 1000 \\
        PNA & 2, 3, 4 & 1 & 32 & $1\mathrm{e}{-3}$ & 32, 64 & 1000 \\
        PDF & 2, 3 & 1 & 32 & $5\mathrm{e}{-4}$ & 64, 129 & 300 \\
        \bottomrule
    \end{tabular}
    \end{adjustbox}
\end{table}

\begin{table}[!htbp]
    \centering
    \caption{Neural hyperparameters used for model selection: optimization and regularization.}
    \label{tab:nn_hyperparameters_optimization}
    \scriptsize
    \renewcommand{\arraystretch}{1.12}
    \setlength{\tabcolsep}{3.2pt}
    \begin{adjustbox}{center,max width=\textwidth}
    \begin{tabular}{lcccccccc}
        \toprule
        \textbf{Model} &
        \textbf{L2} &
        \textbf{Dropout} &
        \makecell{\textbf{Patience}\\\textbf{(loss, acc)}} &
        \textbf{Optimizer} &
        \textbf{Scheduler} &
        \makecell{\textbf{Dense}\\\textbf{dim}} &
        \makecell{\textbf{Embed.}\\\textbf{dim}} &
        \makecell{\textbf{Neighbors}\\\textbf{Aggregation}} \\
        \midrule
        DGCNN & -- & 0.5 & \makecell{500, 500} & Adam & -- & 128 & -- & mean, max, sum \\
        GIN & -- & 0, 0.5 & \makecell{500, 500} & Adam & \makecell{StepLR\\(step 50, $\gamma{=}0.5$)} & -- & -- & sum \\
        GraphSAGE & -- & 0 & \makecell{500, 500} & Adam & -- & -- & -- & mean, max, sum \\
        InfoGraph & \makecell{0, $1\mathrm{e}{-4}$} & 0, 0.1, 0.3 & \makecell{500, 500} & Adam & \makecell{ReduceLROnPlateau\\($\gamma{=}0.5$)} & -- & -- & sum \\
        GNN & -- & 0, 0.5 & \makecell{500, 500} & Adam & \makecell{StepLR\\(step 50, $\gamma{=}0.5$)} & -- & -- & mean, max, sum \\
        GraphCL & -- & 0 & \makecell{500, 500} & Adam & -- & -- & -- & mean, max, sum \\
        FNP & -- & 0.0, 0.2, 0.5 & \makecell{500, 500} & Adam & -- & -- & -- & sum \\
        PNA & -- & 0 & \makecell{500, 500} & Adam & -- & -- & -- & max \\
        PDF & $1\mathrm{e}{-2}$ & 0 & \makecell{500, 500} & Adam & \makecell{StepLR\\(step 50, $\gamma{=}0.5$)} & -- & -- & mean \\
        \bottomrule
    \end{tabular}
    \end{adjustbox}
\end{table}
\begin{table}[H]
    \centering
    \footnotesize  
    \renewcommand{\arraystretch}{1.2}
    \caption{Hyperparameters used in the kernels for model selection in the graph classification task.}
    \label{tab:hyperparameters}
    \begin{adjustbox}{max width=\textwidth}
    \begin{tabular}{l p{3.5cm} p{8.5cm}}
        \toprule
        \textbf{Kernel} & \textbf{Fixed} & \textbf{Validation-tuned} \\
        \midrule
        SM & $k = 3$ & -- \\
        SP & -- & -- \\
        ML & $\gamma = 0.01$, $\eta = 0.01$, $\hat{p} = 10$ & $l_{\max} \in \{0,\dots,5\}$, $\tilde{c} \in \{50,100,200,300\}$ \\
        PK & $w = 10^{-5}$ & $T \in \{1,\dots,6\}$ \\
        HSPPK-WL & Iterations = 20  & $h \in \{0,\dots,5\}$ \\
        HSPPK-SP & Iterations = 20  & $h \in \{0,\dots,5\}$ \\
        GH & -- & Linear kernel / Gaussian kernel \\
        linearFGW-RAW & RBF kernel, $\gamma = 0.1$ & $\alpha \in \{0.1, 0.5, 0.9\}$, GWB layers $= 5$, OT layers $\in \{3,5\}$, Iter $\in \{1,2,3\}$, $\gamma_{\text{kernel}} \in \{0.01, 0.1, 1.0\}$ \\
        linearFGW-WL1 & RBF kernel, $\gamma = 0.1$ & same as above \\
        linearFGW-WL2 & RBF kernel, $\gamma = 0.1$ & same as above \\
        WL (disc) & -- & Iterations $\in \{0,\dots,5\}$ \\
        WLOA & -- & Iterations $\in \{0,\dots,5\}$ \\
        WWL & -- & Iterations $\in \{0,\dots,7\}$, Sinkhorn $\in \{\texttt{False}, \texttt{True}\}$, $\gamma \in \{0.01, 0.1, 1, 10\}$ \\
        SWWL & Projections=16, Quantiles=9, Step=1 & Iterations $\in \{0,\dots,7\}$, $\gamma \in \{0.01, 0.1, 1, 10\}$ \\
        MMD-GK & Bandwidths $= \{1.0, 10.0\}$, $\gamma_{\mathrm{dist}} = 1.0$ & GCN smoothing layers $\in \{1,2,3\}$; SVM $C \in \{10^e: e \in \{-7,-5,-3,-1,0,1,2,3,5,7\}\}$ \\
        NP & -- & Iterations $\in \{0,\dots,5\}$, Linear / Gaussian kernel \\
        NSPDK & $R = 1$, $D = 4$ & -- \\
        NSPDK (disc) & $R = 1$, $D = 4$ & -- \\
        NSPPK & \multicolumn{2}{p{12cm}}{\centering $R = 1$, $D = 4$, $c = 1$, degree cutoff $\tau = 8$, hash bits $b = 16$} \\
        \bottomrule
    \end{tabular}
    \end{adjustbox}
\end{table}

\subsection{Ablation and Sensitivity Experiment Details}
\label{appendix:ablation-details}

Table~\ref{tab:ablation-details} summarizes the classifier and implementation
settings for the ablation in Section~\ref{sec:ablation_sensitivity_density} and
the sensitivity experiments in Appendix~\ref{appendix:sensitivity-results}.
These experiments use explicit NSPPK feature vectors rather than Gram matrices.

\begin{table}[H]
\centering
\caption{Protocol details for the NSPPK ablation and sensitivity experiments.}
\label{tab:ablation-details}
\scriptsize
\setlength{\tabcolsep}{3pt}
\renewcommand{\arraystretch}{1.12}
\begin{adjustbox}{max width=\textwidth}
\begin{tabular}{p{3.0cm}p{3.2cm}p{8.4cm}}
\toprule
\textbf{Experiment} & \textbf{NSPPK setting} & \textbf{Classifier and evaluation details} \\
\midrule
Component ablation &
$R=1$, $D=4$, $c=1$, $b=16$; $\tau=8$ when thresholding is enabled &
Random forest classifier with fixed hyperparameters across
all ablation configurations, random seed 42, and parallel
training. Accuracy is reported over 10 folds. \\
Classifier sensitivity &
$R=1$, $D=4$, $c=1$, $b=11$; no degree cutoff &
Stratified 10-fold outer cross-validation. The evaluated classifiers are
XGBoost, extra-trees, random forest, logistic regression, linear support-vector
classification, and ridge classification. Hyperparameters are selected on the
training portion of each outer fold using the grids shown below. \\
Hyperparameter sensitivity on COX2 &
$R,R'\in\{0,\ldots,6\}$, $D\in\{1,\ldots,6\}$; $b=11$ &
XGBoost classifier with fixed seed 42 and degree cutoff $\tau=3$. The sweep
varies one parameter at a time from the reference setting $(R,D,c)=(1,4,1)$.
\\
Density sensitivity &
$R=1$, $D=4$, $c=1$, $b=11$; no node attributes &
Feature-extraction-only runtime experiment on Erdős--Rényi graphs with
$n=300$ and varying expected average degree. \\
\bottomrule
\end{tabular}
\end{adjustbox}
\end{table}

\begin{table}[H]
\centering
\caption{Classifier grids used in the classifier-sensitivity experiment.}
\label{tab:classifier-sensitivity-grids}
\scriptsize
\setlength{\tabcolsep}{4pt}
\renewcommand{\arraystretch}{1.12}
\begin{adjustbox}{max width=\textwidth}
\begin{tabular}{p{3.4cm}p{10.4cm}}
\toprule
\textbf{Classifier} & \textbf{Grid or fixed configuration} \\
\midrule
XGBoost &
Number of trees $\in\{300,800\}$, maximum depth $\in\{3,6\}$, learning rate
$\in\{0.03,0.1\}$, subsampling rate $\in\{0.8,1.0\}$, histogram tree method,
seed 42, and parallel fitting over all available CPU cores. \\
Extra-trees &
500 trees, maximum depth $\in\{\mathrm{None},4,8\}$, maximum features
$\in\{\sqrt{\cdot},\log_2(\cdot)\}$, seed 42, and parallel fitting over all
available CPU cores. \\
Random forest &
500 trees, maximum depth $\in\{\mathrm{None},4,8\}$, maximum features
$\in\{\sqrt{\cdot},\log_2(\cdot)\}$, seed 42, and parallel fitting over all
available CPU cores. \\
Logistic regression &
$C\in\{0.01,0.1,1,10,100\}$, liblinear solver, maximum 5000 iterations, and
seed 42. \\
Linear support-vector classifier &
$C\in\{0.01,0.1,1,10,100\}$, maximum 10000 iterations, and seed 42. \\
Ridge classifier &
$\alpha\in\{0.1,1,10,100\}$ and seed 42. \\
\bottomrule
\end{tabular}
\end{adjustbox}
\end{table}

\subsection{Robustness Study Hyperparameters}
\label{appendix:robustness}

Table~\ref{tab:robustness-hparams} summarizes the configurations used for the diagonal dominance / robustness experiments.
Neural baselines (GIN-Random, GraphCL, InfoGraph) were run with a common lightweight setup, while classical kernels (GraphHopper, Propagation Kernel) followed their standard definitions. NSPPK used $b=12$, with the same structural radii and degree cutoff as in the main experiments.
For every method, we normalize the Gram matrix as
$\widehat K_{ij}=K_{ij}/\sqrt{K_{ii}K_{jj}}$ before constructing the similarity
curves, heatmaps, and singular-spectrum entropy. The heatmaps provide the direct
diagnostic of off-diagonal coupling; entropy is reported separately as an
effective-rank diagnostic.

\begin{table}[H]
\centering
\caption{Hyperparameters for robustness / diagonal dominance analysis.}
\label{tab:robustness-hparams}
\renewcommand{\arraystretch}{1.1}
\setlength{\tabcolsep}{6pt}
\begin{tabular}{p{3.2cm} p{9.5cm}}
\toprule
\textbf{Method} & \textbf{Configuration} \\
\midrule
\textbf{NSPPK} & $R=1$, $D=4$, $c=1$, $\tau=8$, $b=12$ \\
\textbf{GIN-Random} & 3 layers GIN, hidden dim=32, MLP layers=2, pooling=sum, epochs=200, lr=0.01, seed=42, orthogonal init, no supervision \\
\textbf{GraphCL} & Same GIN backbone as above, contrastive pretraining with augmentations, 200 epochs, lr=0.01 \\
\textbf{InfoGraph} & Same GIN backbone as above, maximizing mutual information, 200 epochs, lr=0.01 \\
\textbf{GraphHopper} & Shortest-path kernel, weight decay $w=10^{-5}$, $t_{\max} \in \{1,2,3,4,5\}$ \\
\textbf{Propagation Kernel} & Attribute propagation with $M=L1$ distance, 5 iterations \\
\bottomrule
\end{tabular}
\end{table}

The neural robustness setup includes 50{,}000 graphs with 50 to 250 nodes.
GraphCL and InfoGraph use their respective pretraining objectives.

\section{Hash-width experiment on QM9}
\label{sec:qm9}

Our preprocessed collection contains about 112{,}000 of the 133{,}885 QM9
molecules~\citep{ramakrishnan2014quantum}, represented as graphs with continuous
node attributes. Vectorization took 282--312 seconds at 2--11 bits and
2{,}285--2{,}958 seconds at 12--14 bits, when we observed memory swapping
(Figure~\ref{fig:nsppk_qm9_main}). The 15-bit run failed during memory allocation.

\section{Scalability of NSPPK Vectorization}
\label{appendix:scaling}

\begin{figure}[htbp]
  \centering
  \includegraphics[width=0.54\linewidth]{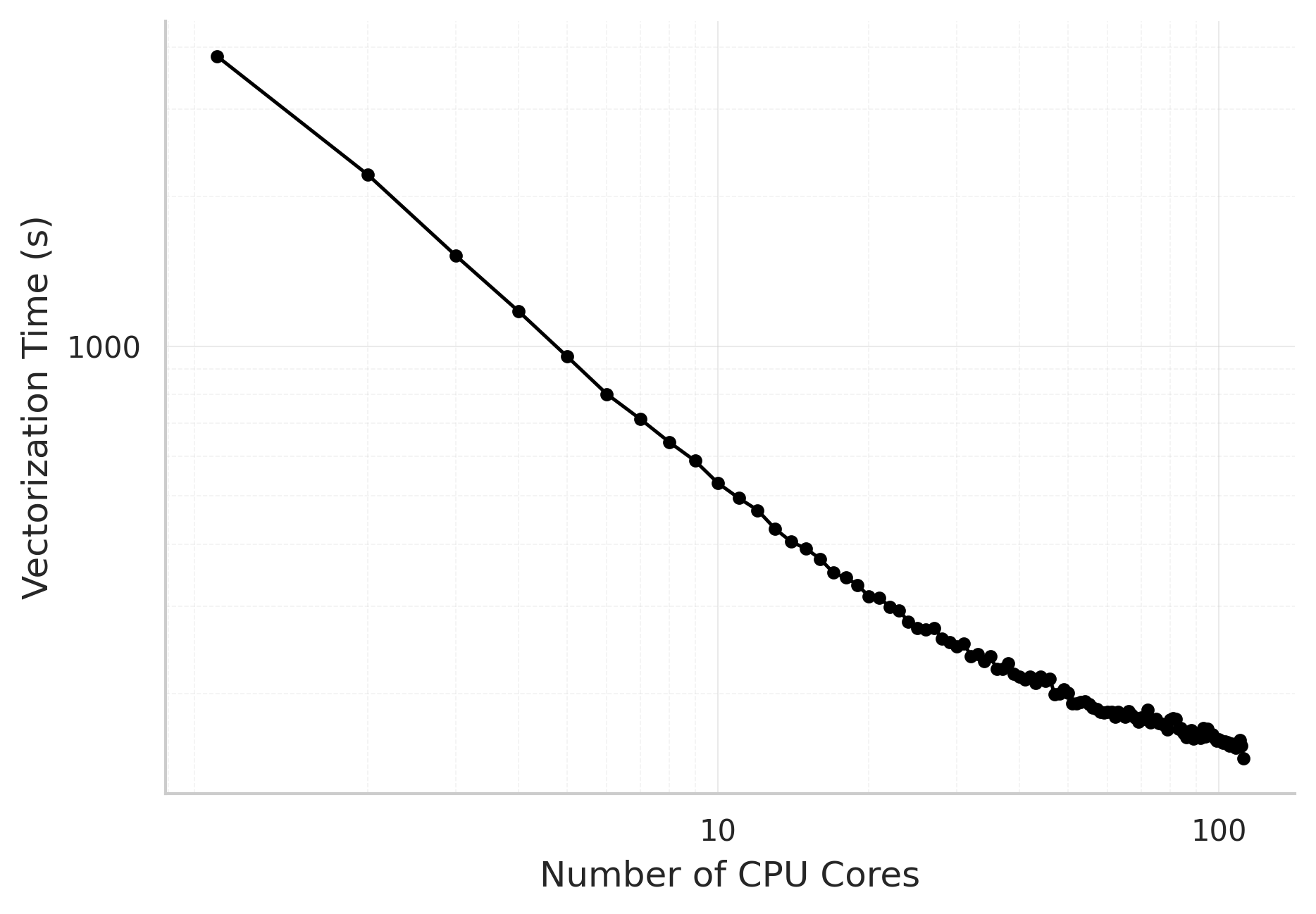}
  \caption{Vectorization time by logical CPU core count on the preprocessed
  QM9 collection (log--log scale).}
  \label{fig:appendix_scaling_plot}
\end{figure}

We used 12-bit hashing with $R=1$, $D=4$, and $c=1$ for the preprocessed
QM9 collection of about 112{,}000 graphs. The machine had two Intel Xeon Gold
6330 processors with 28 physical cores each, for 112 logical CPUs across two
NUMA nodes. The log--log plot in Figure~\ref{fig:appendix_scaling_plot} shows diminishing
reductions in wall-clock time as the number of workers increases. At 112
logical CPUs, vectorization takes under 10 minutes.

\section{Additional Experiments on Citation Networks}
\label{appendix:citation}

To evaluate NSPPK outside the molecular benchmarks, we additionally tested it
on three widely used citation network datasets with
fundamentally different structural properties: \textbf{Cora},
\textbf{CiteSeer}, and \textbf{PubMed}. These graphs exhibit higher-degree
outliers, broader graph diameters, and citation-style connectivity patterns,
in contrast to the small and sparse molecular graphs considered elsewhere
in this paper.

\begin{table}[htbp]
\centering
\caption{Citation network sizes and NSPPK encoding times (80/20 split, seed 42).}
\label{tab:citation_runtime}
\scriptsize
\setlength{\tabcolsep}{4pt}
\begin{tabular}{lrrrrr}
\toprule
\textbf{Dataset} &
\textbf{\#Nodes} &
\textbf{\#Edges} &
\textbf{Avg Deg.} &
\textbf{Max Deg.} &

\textbf{Encoding Time (s)} \\
\midrule
Cora     & 2{,}708  & 5{,}278  & 3.90 & 168 & 0.7897 \\
CiteSeer & 3{,}327  & 4{,}552  & 2.74 & 99  & 0.7447 \\
PubMed   & 19{,}717 & 44{,}324 & 4.50 & 171 & 0.8747  \\
\bottomrule
\end{tabular}
\end{table}

We compared NSPPK against a  set of commonly used graph neural network
architectures for node classification: GCN~\citep{kipf2017semi}, GAT~\citep{velivckovic2018gat}, GraphSAGE~\citep{hamilton2017graphsage}, GIN~\citep{xu2019powerful}, SGC~\citep{wu2019simplifying}, APPNP~\citep{klicpera2019predict},
and a feature-only MLP.

All GNN models were trained directly on the original citation graphs using
the same 80/20 train--test split as NSPPK.
For GCN, GIN, GraphSAGE, and GAT, we used two-layer architectures with a hidden
dimension of 64, ReLU activations, and dropout applied between layers.
GAT employed eight attention heads in the first layer.
APPNP was implemented as a two-layer MLP followed by personalized PageRank
propagation with $K{=}10$ steps and teleport probability $\alpha{=}0.1$.
SGC used a single linear classifier with $K{=}2$ propagation steps.
The MLP baseline consisted of a two-layer feed-forward network operating
solely on node features, without using graph structure.

All models were trained using the Adam optimizer
(learning rate 0.01, weight decay $5\times10^{-4}$)
for 200 epochs.
Model selection relied solely on the fixed training split, without tuning
hyperparameters per dataset.
Performance is reported on the held-out test nodes in terms of
macro-averaged ROC--AUC (one-vs-rest).

For NSPPK, we followed a standard node classification setup with an 80/20 train--test
split and a fixed random seed of 42. All three citation networks used a common
configuration: $R=1$, $D=4$, $c=0$, and 11-bit hashing.
Continuous node attributes were projected to 24 dimensions using SVD.
Discrete node labels were obtained by applying $k$-means
clustering with $k=5$ to the node attributes. Continuous
node attributes were separately reduced to 24 dimensions
using singular value decomposition (SVD).
\begin{table}[htbp]
\centering
\caption{Test ROC--AUC (macro, one-vs-rest) on citation networks with an 80/20 node classification split.
}
\label{tab:citation_auc}
\setlength{\tabcolsep}{4pt}
\begin{tabular}{lcccccccc}

\toprule
Dataset
& GraphSAGE
& GAT
& APPNP
& GCN
& SGC
& NSPPK+XGB
& MLP
& GIN \\
\midrule
CiteSeer
& 0.9422
& 0.9454
& 0.9310
& 0.9257
& 0.9432
& 0.9253
& 0.9247
& 0.9084 \\

Cora
& 0.9846
& 0.9874
& 0.9884
& 0.9840
& 0.9874
& 0.9636
& 0.9502
& 0.9669 \\

PubMed
& 0.9759
& 0.9615
& 0.9622
& 0.9659
& 0.9396
& 0.9686
& 0.9714
& 0.9663 \\
\bottomrule
\end{tabular}
\end{table}

NSPPK with XGBoost attains macro ROC--AUC of 0.9253 on CiteSeer, 0.9636 on
Cora, and 0.9686 on PubMed. GraphSAGE records the highest PubMed score (0.9759).

\section{Further Visualization: Accuracy–Time Trade-off}
\label{appendix:tradeoff}

\begin{figure}[!htb]
\centering
\captionsetup{font=small,aboveskip=3pt,belowskip=0pt}
\captionsetup[subfigure]{font=small}

\begin{subfigure}[t]{0.32\textwidth}
  \centering\includegraphics[width=\linewidth]{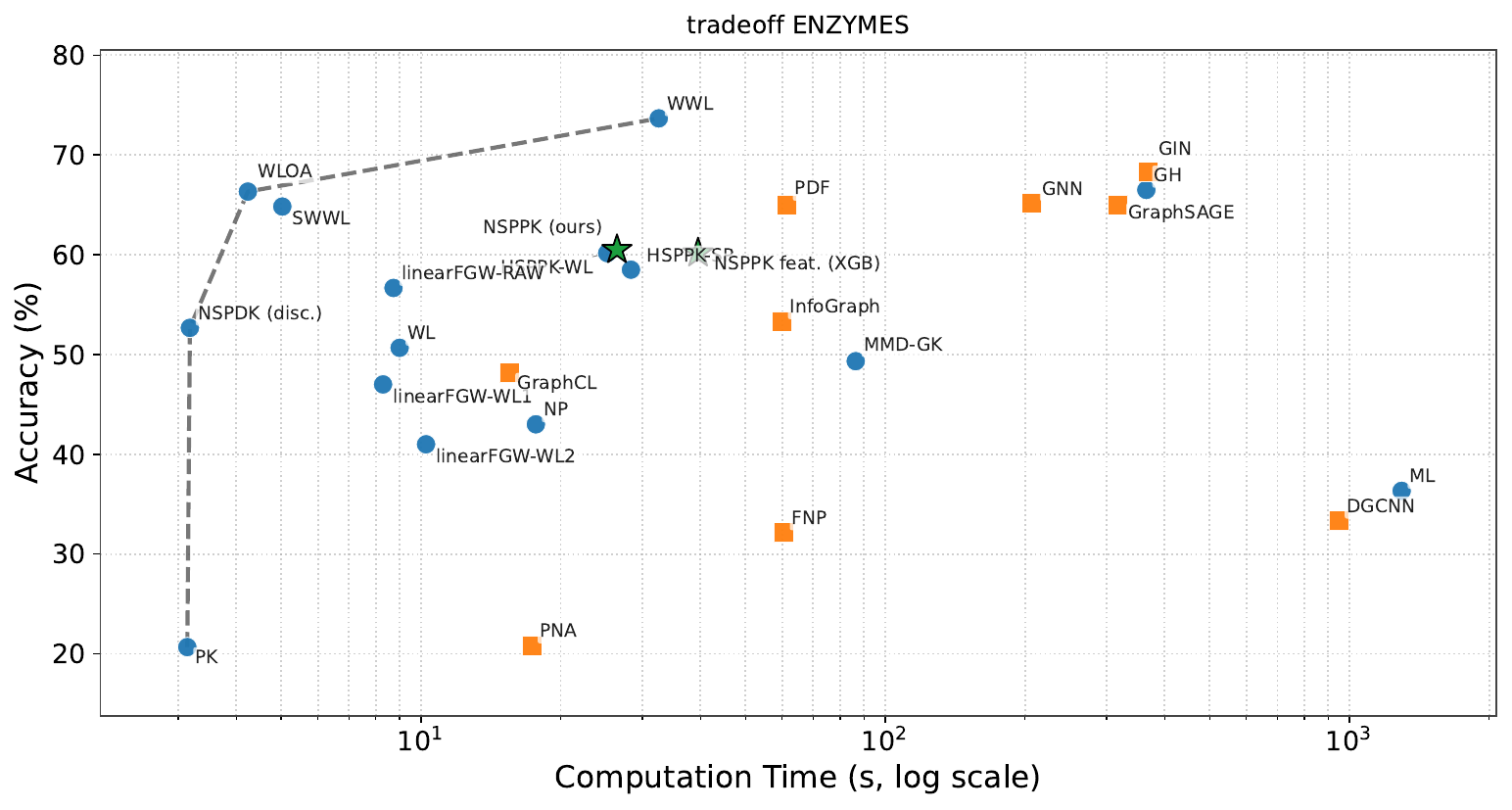}
  \caption{ENZYMES}
\end{subfigure}\hfill
\begin{subfigure}[t]{0.32\textwidth}
  \centering\includegraphics[width=\linewidth]{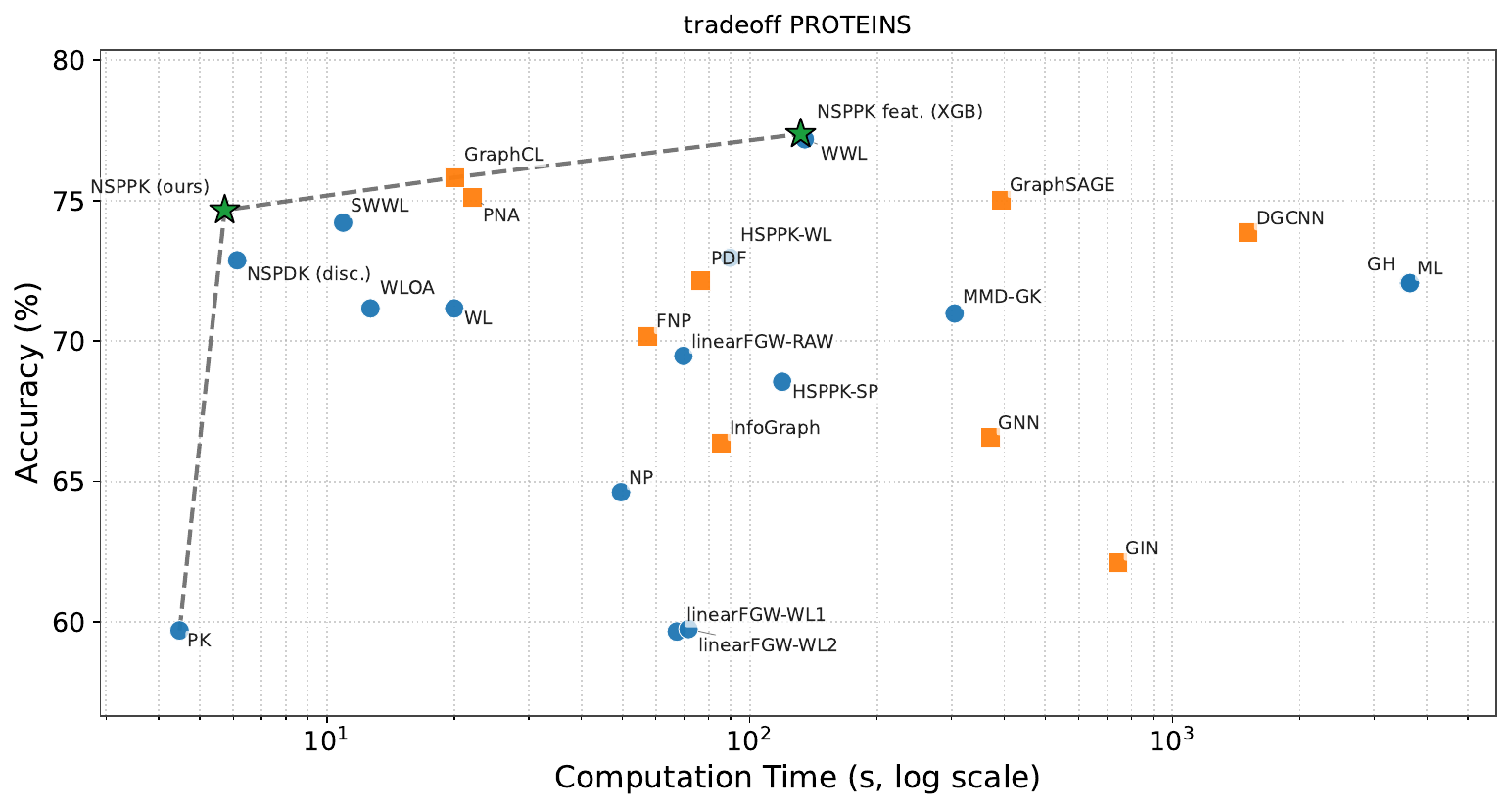}
  \caption{PROTEINS\_full}
\end{subfigure}\hfill
\begin{subfigure}[t]{0.32\textwidth}
  \centering\includegraphics[width=\linewidth]{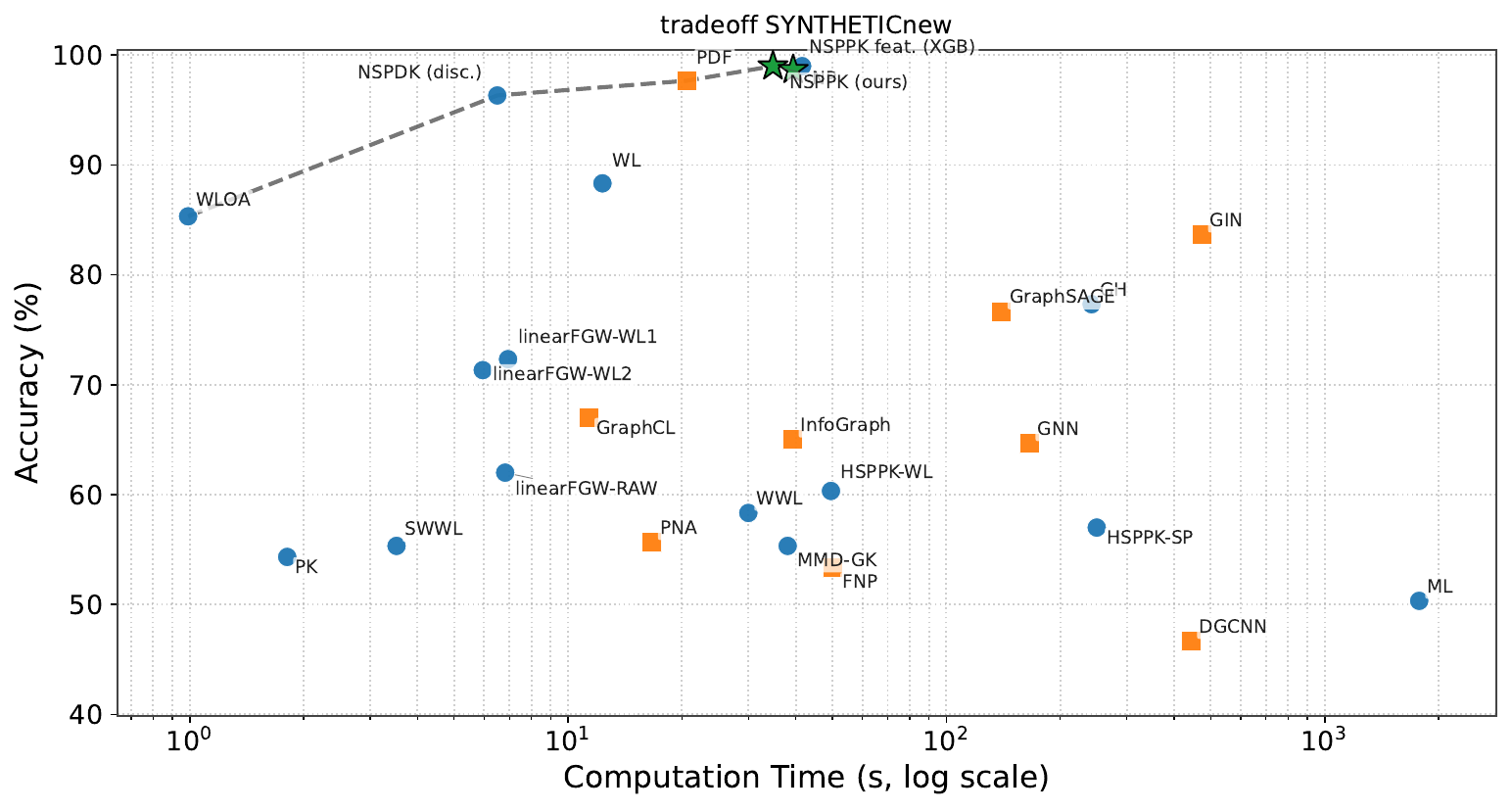}
  \caption{SYNTHETICnew}
\end{subfigure}

\vspace{2pt}

\begin{subfigure}[t]{0.32\textwidth}
  \centering\includegraphics[width=\linewidth]{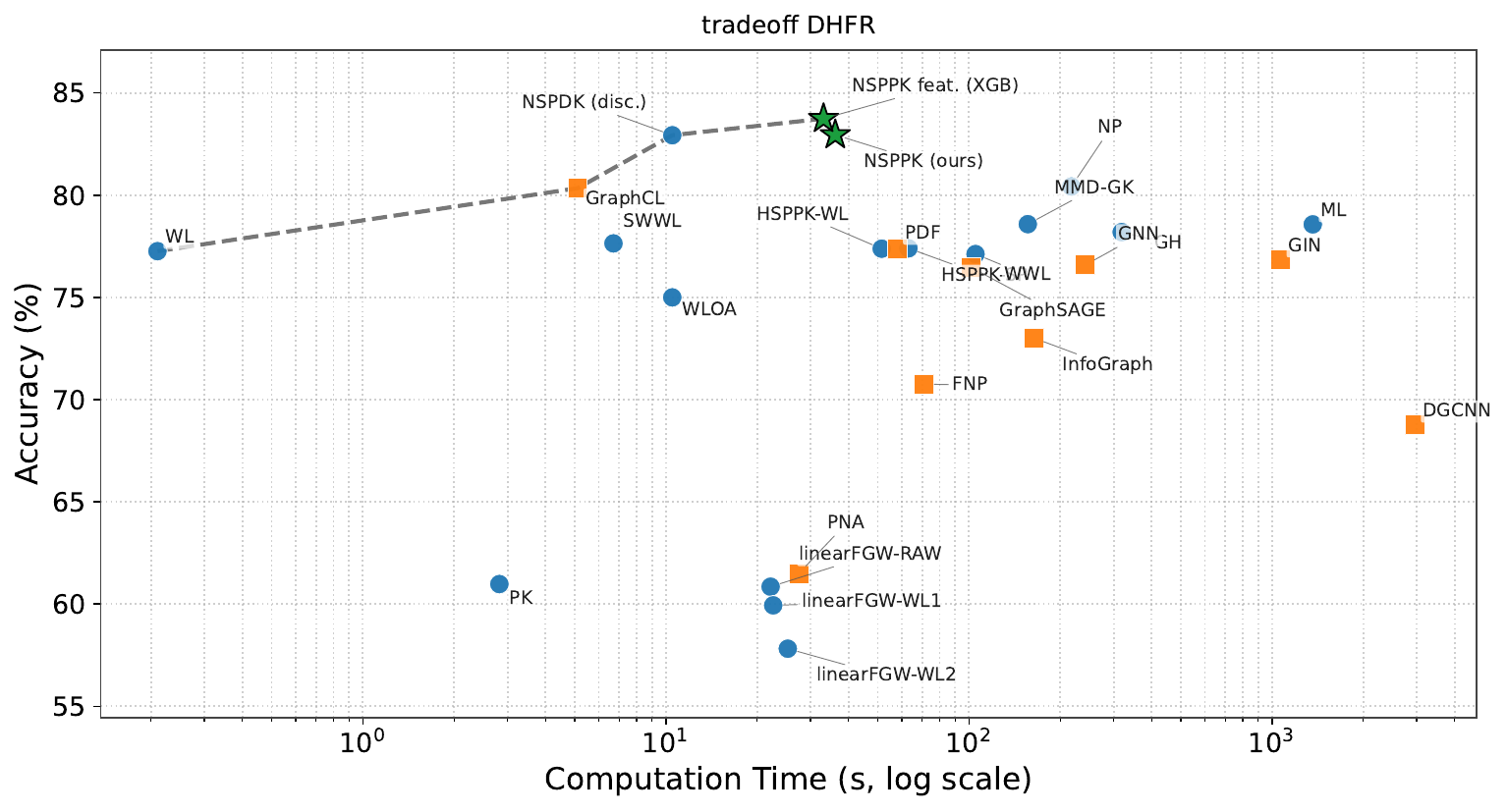}
  \caption{DHFR}
\end{subfigure}\hfill
\begin{subfigure}[t]{0.32\textwidth}
  \centering\includegraphics[width=\linewidth]{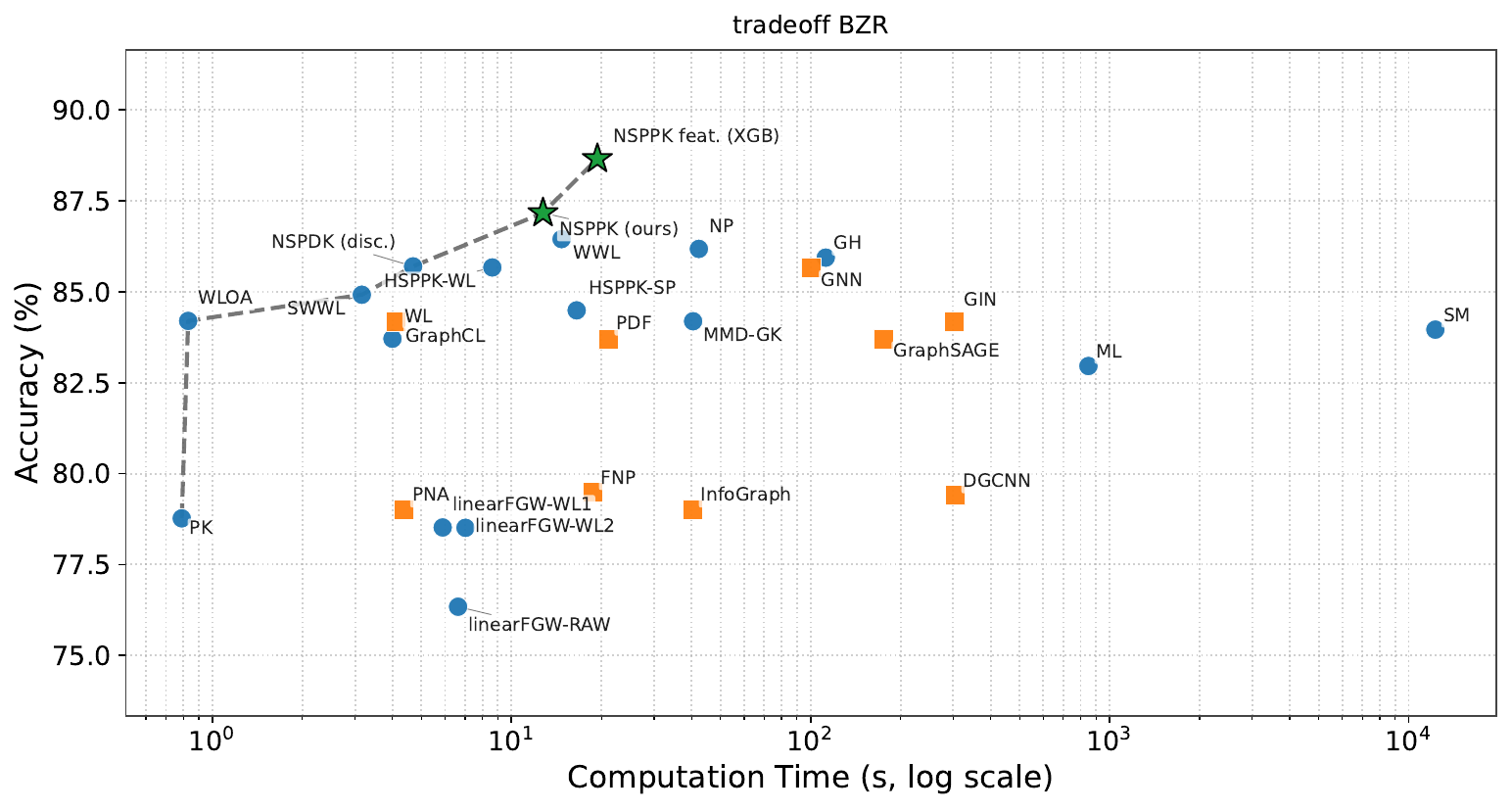}
  \caption{BZR}
\end{subfigure}\hfill
\begin{subfigure}[t]{0.32\textwidth}
  \centering\includegraphics[width=\linewidth]{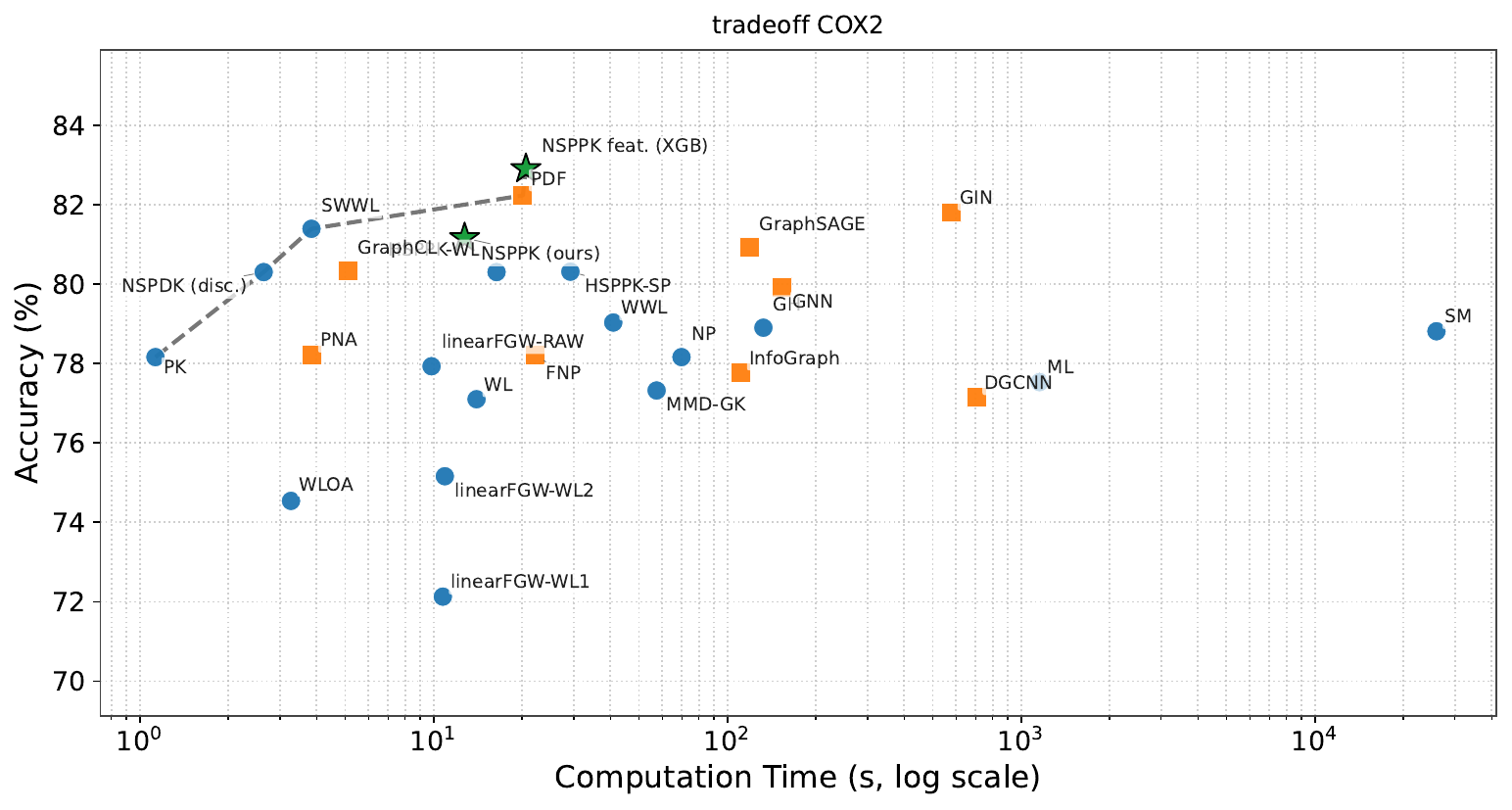}
  \caption{COX2}
\end{subfigure}

\vspace{3pt}

\begin{subfigure}[t]{0.98\textwidth}
  \centering\includegraphics[width=\linewidth]{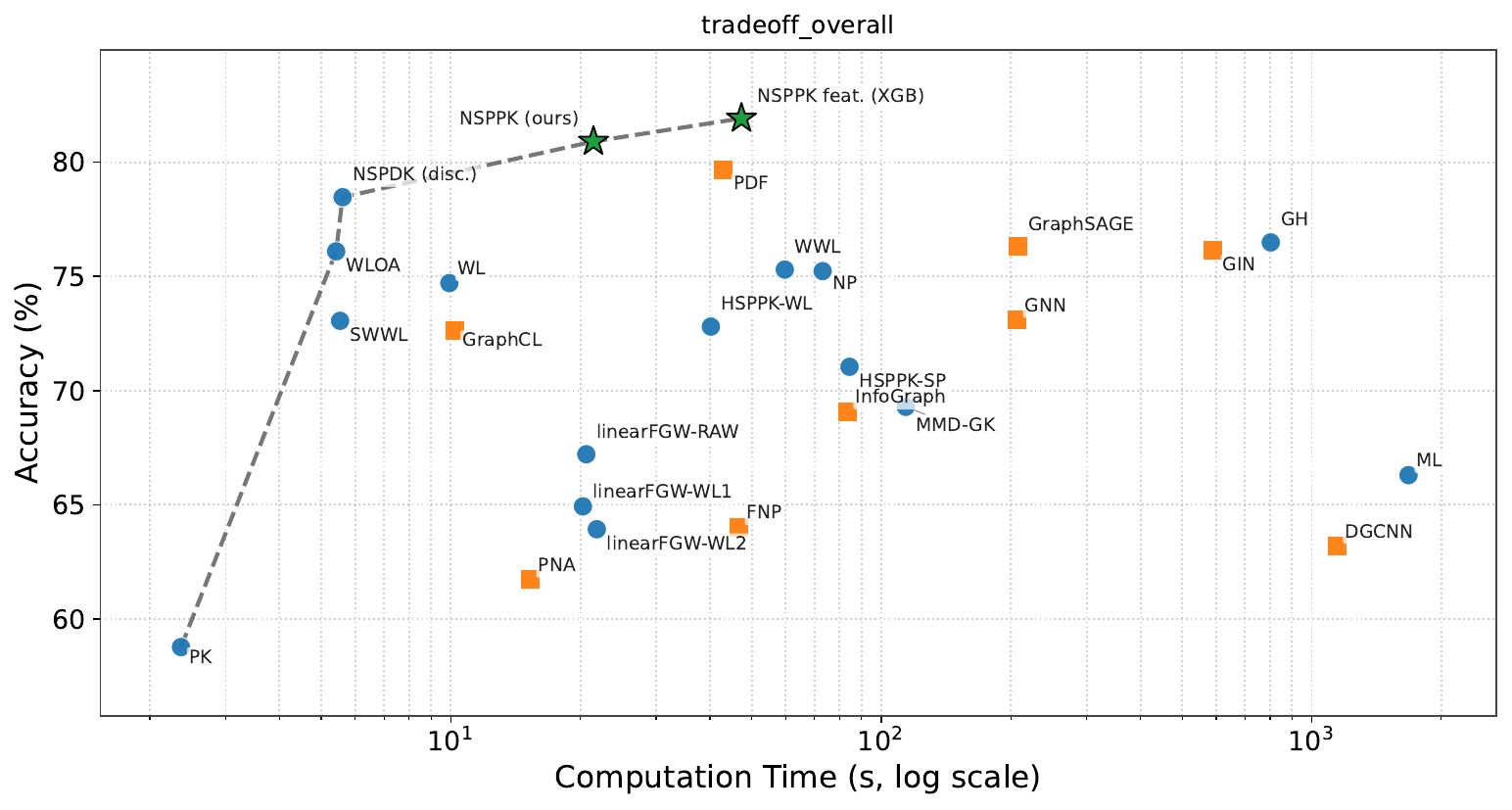}
  \caption{Overall (average across datasets)}
\end{subfigure}

\caption{\textbf{Accuracy–time trade-off (log time).}
Markers: green star = NSPPK, blue circle = other kernels, orange square = neural nets.
Dashed line = Pareto front.}
\label{fig:tradeoff_all}
\end{figure}

The aggregate panel in Figure~\ref{fig:tradeoff_all} places NSPPK on the plotted
frontier. The kernel and neural runtimes in this figure use the hardware
settings described in Section~\ref{sec:small_datasets}.

\FloatBarrier
\subsection{Statistical comparison of accuracy ranks}
\label{appendix:cd}

We apply a Friedman test to the mean accuracies displayed in
Tables~\ref{tab:with_attr} and~\ref{tab:nn_with_attr}, treating each of the six
datasets as a block and averaging tied ranks within each block. The analysis
excludes the attributes-only control and any method that timed out on at least
one dataset. The omnibus $p$-values are $3.35\times10^{-5}$ for the 16 kernel
methods and $2.16\times10^{-4}$ for the 10 pipelines in the neural-baseline
comparison.

\begin{figure}[H]
  \centering
  \begin{subfigure}{0.97\linewidth}
    \centering
    \includegraphics[width=\linewidth]{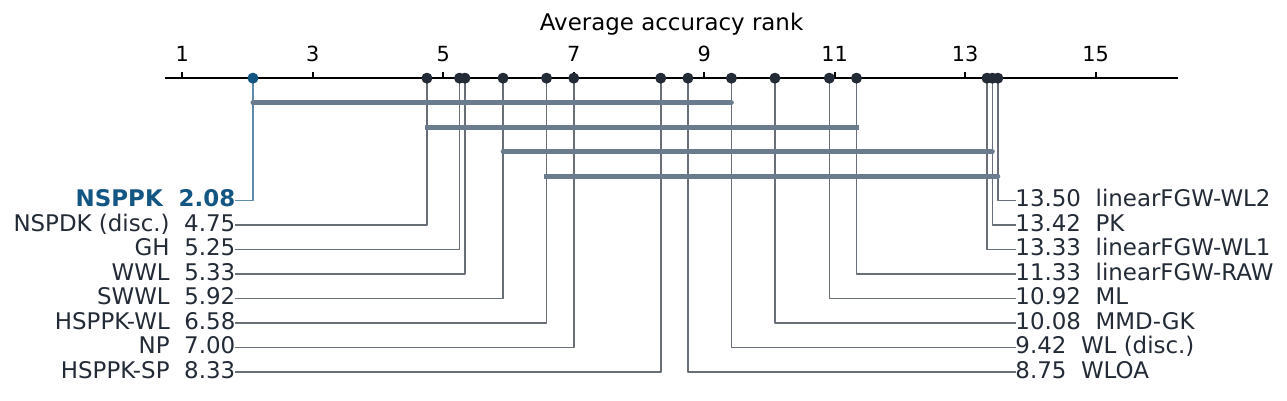}
    \caption{Graph kernels.}
  \end{subfigure}\par\medskip
  \begin{subfigure}{0.97\linewidth}
    \centering
    \includegraphics[width=\linewidth]{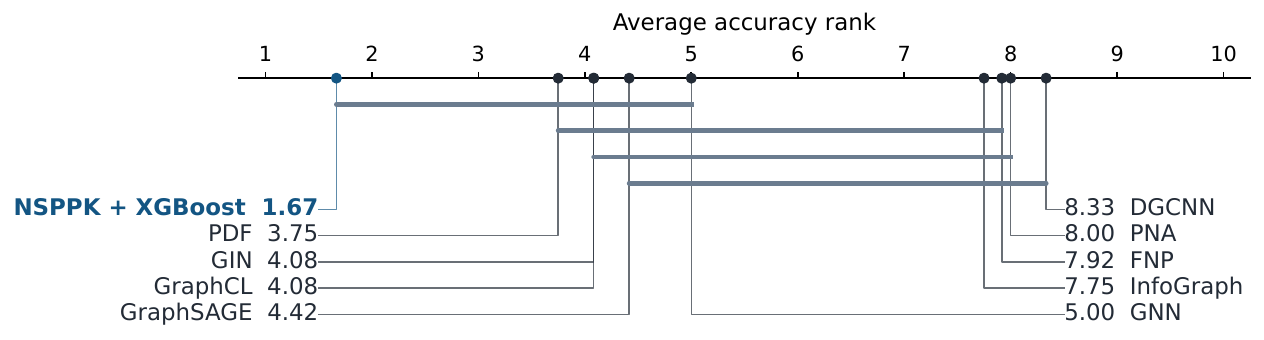}
    \caption{Neural baselines and NSPPK with XGBoost.}
  \end{subfigure}
  \caption{Average accuracy ranks across six datasets. Lower ranks are better.
  A crossbar links a group with no significant pairwise differences at the
  $0.05$ level after Holm adjustment of the Conover tests.}
  \label{fig:accuracy-ranks}
\end{figure}

We follow each omnibus test with Conover pairwise comparisons and apply Holm
adjustment across all pairs within that comparison. At the $0.05$ level,
NSPPK ranks significantly ahead of ML, PK, the three linearFGW variants,
and MMD-GK in the kernel comparison. With XGBoost, it ranks significantly
ahead of DGCNN, InfoGraph, FNP, and PNA in the neural-baseline comparison.
Figure~\ref{fig:accuracy-ranks} shows the average ranks and the groups from
the adjusted pairwise tests.

\section{Additional Results: No-Attribute Setting}
\label{appendix:no_attr}

Tables~\ref{tab:no_attr} and~\ref{tab:nn_no_attr} report kernel and neural network accuracy, respectively, when node attributes are removed.
This comparison isolates the contribution of structural features under the
reported protocol. NSPPK has an average rank of \(5.00\) in the kernel
comparison and \(3.33\) in the comparison with neural baselines.

\begin{table}[!htb]
\centering
\caption{Classification accuracy (\%) \textbf{without} node attributes (-) with Avg Rank.}
\label{tab:no_attr}
\scriptsize
\setlength{\tabcolsep}{3.2pt}
\begin{adjustbox}{max width=\textwidth}
\begin{tabular}{lccccccc}
\toprule
\textbf{Method} & \textbf{SYNTHETICnew} & \textbf{DHFR} & \textbf{BZR} & \textbf{COX2} & \textbf{ENZYMES} & \textbf{PROTEINS\_full} & \textbf{Avg Rank} \\
\midrule
SM  & \TO & \TO & \mps{79.02}{1.10} & \mps{78.16}{0.81} & \TO & \TO & 15.50 \\
SP  & \TO & \TO & \TO & \TO & \TO & \TO & 18.67 \\
ML  & \mps{67.22}{9.51} & \mps{79.90}{1.36} & \mps{86.63}{3.81} & \mps{77.95}{4.63} & \mps{33.66}{5.31} & \mps{72.80}{3.80} & 5.67 \\
PK  & \mps{61.33}{7.33} & \mps{60.98}{0.48} & \mps{78.77}{1.01} & \mps{78.16}{8.07} & \mps{18.33}{5.22} & \mps{58.48}{4.29} & 12.17 \\
HSPPK\_WL & \mps{50.00}{0.00} & \mps{57.56}{6.67} & \mps{80.26}{3.02} & \mps{72.41}{17.06} & \mps{16.50}{2.73} & \mps{63.80}{5.76} & 15.17 \\
HSPPK\_SP & \mps{58.00}{7.18} & \mps{64.42}{4.05} & \mps{76.26}{9.39} & \mps{77.31}{4.70} & \mps{21.00}{5.92} & \mps{46.20}{4.70} & 13.17 \\
GH  & \mps{59.33}{9.28} & \mps{74.08}{4.27} & \mps{81.25}{2.40} & \mps{77.30}{3.14} & \mps{25.17}{3.98} & \mps{71.61}{4.32} & 8.50 \\
NP  & \mps{97.00}{3.15} & \mps{80.43}{4.01} & \mps{84.16}{5.65} & \mps{80.29}{3.42} & \mps{36.70}{0.53} & \mps{69.99}{3.88} & 4.50 \\
linearFGW-RAW & \mps{57.33}{6.29} & \mps{54.90}{4.77} & \mps{80.52}{3.76} & \mps{76.24}{4.74} & \mps{23.00}{4.88} & \mps{71.35}{4.56} & 11.67 \\
linearFGW-WL1 & \mps{56.00}{11.72} & \mps{63.90}{3.97} & \mps{80.99}{5.24} & \mps{76.45}{1.86} & \mps{24.50}{4.78} & \mps{70.17}{4.81} & 11.00 \\
linearFGW-WL2 & \mps{48.67}{7.33} & \mps{59.12}{6.41} & \mps{79.48}{5.48} & \mps{74.74}{5.12} & \mps{22.33}{5.59} & \mps{69.54}{3.12} & 14.00 \\
WL (disc.) & \mps{79.00}{12.39} & \mps{80.56}{5.29} & \mps{87.90}{3.92} & \mps{78.17}{3.48} & \mps{40.17}{7.54} & \mps{69.00}{4.08} & 4.67 \\
WLOA & \mps{81.00}{6.16} & \mps{80.96}{3.90} & \mps{83.71}{8.36} & \mps{78.16}{2.75} & \mps{42.67}{4.84} & \mps{74.49}{3.53} & \textbf{4.00} \\
WWL & \mps{50.00}{0.00} & \mps{60.98}{0.47} & \mps{78.77}{1.01} & \mps{78.16}{0.80} & \mps{16.67}{0.00} & \mps{59.57}{0.17} & 13.17 \\
SWWL & \mps{50.0}{0.00} & \mps{60.98}{0.48} & \mps{78.77}{1.01} & \mps{78.16}{0.80} & \mps{16.67}{0.00} & \mps{59.57}{0.17} & 13.17 \\
NSPDK & \mps{95.33}{3.72} & \mps{82.94}{5.70} & \mps{85.68}{4.04} & \mps{77.09}{3.84} & \mps{35.67}{9.22} & \mps{71.33}{3.06} & 5.50 \\
NSPDK (disc.) & \mps{95.33}{3.72} & \mps{82.94}{5.70} & \mps{85.68}{4.04} & \mps{77.09}{3.84} & \mps{35.67}{9.22} & \mps{71.33}{3.06} & 5.50 \\
MMD-GK & \mps{84.33}{6.16} & \mps{75.66}{5.48} & \mps{84.44}{3.98} & \mps{76.03}{3.94} & \mps{30.33}{4.46} & \mps{69.09}{3.01} & 9.00 \\
\rowcolor{gray!10}
\textbf{NSPPK (ours)} & \mps{98.00}{3.93} & \mps{80.18}{5.58} & \mps{87.65}{4.54} & \mps{77.73}{3.96} & \mps{33.17}{5.80} & \mps{71.34}{4.06} & 5.00 \\

\bottomrule
\end{tabular}
\end{adjustbox}

\vspace{0.5ex}
\footnotesize\emph{Note:} TIMEOUT denotes a run exceeding 24 hours for one fold and is assigned the tied worst rank on that dataset. Avg Rank is computed across all six datasets; lower is better.
\end{table}

\begin{table}[!htb]
\centering
\caption{Classification accuracy (\%) without node attributes in the neural-baseline
comparison. Ranks use the displayed mean accuracies, with ties assigned the
average rank.}
\label{tab:nn_no_attr}
\scriptsize
\setlength{\tabcolsep}{3.2pt}
\begin{adjustbox}{max width=\textwidth}
\begin{tabular}{lccccccc}
\toprule
\textbf{Method} &
\textbf{SYNTHETICnew} &
\textbf{DHFR} &
\textbf{BZR} &
\textbf{COX2} &
\textbf{ENZYMES} &
\textbf{PROTEINS\_full} &
\textbf{Avg Rank} \\
\midrule
DGCNN     & \mps{44.67}{6.86} & \mps{61.50}{4.93} & \mps{81.98}{2.20} & \mps{78.22}{0.07} & \mps{26.80}{7.09} & \mps{73.22}{3.48} & 6.83 \\
GraphSAGE & \mps{43.33}{5.58} & \mps{78.06}{6.25} & \mps{83.70}{5.59} & \mps{80.30}{0.03} & \mps{48.17}{7.58} & \mps{74.93}{2.82} & 4.92 \\
InfoGraph & \mps{67.33}{21.54} & \mps{71.53}{8.51} & \mps{75.05}{15.04} & \mps{69.02}{0.20} & \mps{53.33}{4.79} & \mps{63.07}{4.69} & 6.50 \\
GIN       & \mps{53.00}{9.71} & \mps{79.91}{7.91} & \mps{73.95}{3.30} & \mps{79.91}{0.08} & \mps{42.67}{7.68} & \mps{65.77}{5.02} & 6.00 \\
GraphCL   & \mps{50.00}{8.69} & \mps{80.16}{2.51} & \mps{79.99}{3.62} & \mps{81.38}{3.79} & \mps{37.50}{5.12} & \mps{71.43}{3.92} & 5.25 \\
GNN       & \mps{43.33}{5.58} & \mps{82.67}{4.96} & \mps{84.66}{4.60} & \mps{81.60}{5.54} & \mps{48.50}{5.80} & \mps{71.79}{3.61} & 3.75 \\
FNP       & \mps{50.00}{8.69} & \mps{60.85}{4.63} & \mps{81.73}{3.52} & \mps{78.22}{3.94} & \mps{35.83}{7.79} & \mps{72.41}{3.70} & 6.83 \\
PNA       & \mps{46.00}{7.72} & \mps{60.98}{4.46} & \mps{78.76}{4.33} & \mps{78.22}{7.01} & \mps{18.83}{7.07} & \mps{70.62}{3.69} & 8.33 \\
PDF       & \mps{50.00}{8.69} & \mps{78.83}{4.21} & \mps{84.43}{4.63} & \mps{81.38}{3.79} & \mps{52.00}{5.26} & \mps{74.75}{2.28} & 3.25 \\
\rowcolor{gray!10}
NSPPK feat.\ (XGBoost)
          & \mps{91.00}{4.73} & \mps{78.44}{1.27} & \mps{89.60}{3.29} & \mps{82.76}{4.25} & \mps{41.83}{5.55} & \mps{71.60}{3.15} & \textbf{3.33} \\
\bottomrule
\end{tabular}
\end{adjustbox}
\end{table}

\section{Additional Runtimes: No-Attribute Setting}
\label{appendix:no_attr_runtime}

Tables~\ref{tab:no_attr_runtime_kernels} and~\ref{tab:no_attr_runtime_nns} report computation times for kernels and neural networks without node attributes.
In this setting, NSPPK's average runtime rank is reported alongside its
predictive rank so that the runtime--accuracy trade-off can be assessed from the
two tables without assuming that either ranking transfers to other datasets.

\begin{table}[!htb]
\centering
\caption{Runtimes (seconds) without node attributes in the neural-baseline
comparison. The final columns average the displayed times and their
within-dataset ranks. Faster runs receive lower ranks.}
\label{tab:no_attr_runtime_nns}
\scriptsize
\setlength{\tabcolsep}{3.5pt}
\begin{adjustbox}{max width=\textwidth}
\begin{tabular}{lcccccccc}
\toprule
\textbf{Method} & \textbf{SYNTHETICnew} & \textbf{DHFR} & \textbf{BZR} & \textbf{COX2} & \textbf{ENZYMES} & \textbf{PROTEINS\_full} & \textbf{Average} & \textbf{Rank}\\
\midrule
DGCNN & 428.69\,s & 575.14\,s & 571.26\,s & 237.85\,s & 887.36\,s & 1611.41\,s & 718.62\,s & 9.83\\
GraphSAGE & 124.81\,s & 109.79\,s & 126.21\,s & 140.69\,s & 248.03\,s & 395.38\,s & 190.82\,s & 7.50\\
InfoGraph & 65.27\,s & 185.02\,s & 100.00\,s & 87.92\,s & 144.01\,s & 241.47\,s & 137.28\,s & 6.33\\
GIN & 388.95\,s & 392.46\,s & 358.57\,s & 392.46\,s & 440.30\,s & 884.01\,s & 476.12\,s & 9.17\\
GraphCL & 3.58\,s & 13.53\,s & 4.23\,s & 5.42\,s & 13.28\,s & 17.61\,s & 9.61\,s & 2.00\\
GNN & 81.93\,s & 255.75\,s & 28.95\,s & 122.86\,s & 247.42\,s & 403.82\,s & 190.12\,s & 7.00\\
FNP & 7.69\,s & 92.75\,s & 63.45\,s & 15.82\,s & 70.24\,s & 38.84\,s & 48.13\,s & 4.33\\
PNA & 3.32\,s & 19.04\,s & 3.83\,s & 2.80\,s & 8.59\,s & 23.63\,s & 10.20\,s & \textbf{1.67}\\
PDF & 13.12\,s & 75.36\,s & 17.80\,s & 23.45\,s & 63.85\,s & 67.34\,s & 43.49\,s & 4.33\\
NSPPK feat.\ (XGB) & 32.91\,s & 10.07\,s & 4.28\,s & 6.25\,s & 7.43\,s & 60.64\,s & 20.26\,s & 2.83\\
\bottomrule
\end{tabular}
\end{adjustbox}
\end{table}

\begin{table}[!htb]
\centering
\caption{Runtime (seconds) \textbf{without} node attributes (\textbf{lower is better}).}
\label{tab:no_attr_runtime_kernels}

\scriptsize
\setlength{\tabcolsep}{3.5pt}
\begin{adjustbox}{max width=\textwidth}
\begin{tabular}{lcccccccc}
\toprule
\textbf{Method} & \textbf{SYNTHETICnew} & \textbf{DHFR} & \textbf{BZR} & \textbf{COX2} & \textbf{ENZYMES} & \textbf{PROTEINS\_full} & \textbf{Average} & \textbf{Rank}\\
\midrule
SM & \TO & \TO & 11853.30\,s & 25478.50\,s & \TO & \TO & 18665.90\,s & 18.00\\
SP & \TO & \TO & \TO & \TO & \TO & \TO & -- & --\\
ML & 978.86\,s & 1300.05\,s & 158.40\,s & 1509.62\,s & 1792.43\,s & 5661.68\,s & 1900.17\,s & 17.00\\
PK & 0.59\,s & 0.66\,s & 0.23\,s & 0.33\,s & 0.78\,s & 2.11\,s & 0.78\,s & 2.00\\
HSPPK\_WL & 38.74\,s & 29.97\,s & 8.76\,s & 11.30\,s & 19.83\,s & 76.69\,s & 30.88\,s & 11.00\\
HSPPK\_SP & 285.63\,s & 137.62\,s & 18.33\,s & 25.21\,s & 31.37\,s & 121.37\,s & 103.26\,s & 13.33\\
GH & 178.47\,s & 322.97\,s & 99.82\,s & 132.87\,s & 365.25\,s & 791.82\,s & 315.20\,s & 15.83\\
NP & 66.93\,s & 279.77\,s & 44.90\,s & 97.51\,s & 69.69\,s & 240.32\,s & 133.19\,s & 14.50\\
linearFGW-RAW & 6.21\,s & 19.98\,s & 6.17\,s & 8.75\,s & 12.91\,s & 76.17\,s & 21.70\,s & 8.67\\
linearFGW-WL1 & 7.00\,s & 20.01\,s & 6.24\,s & 10.22\,s & 8.51\,s & 52.96\,s & 17.49\,s & 8.67\\
linearFGW-WL2 & 6.17\,s & 25.21\,s & 5.21\,s & 11.27\,s & 11.82\,s & 74.40\,s & 22.35\,s & 8.50\\
WL (disc.) & 0.14\,s & 0.25\,s & 0.05\,s & 0.12\,s & 0.13\,s & 0.34\,s & 0.17\,s & \textbf{1.00}\\
WLOA & 0.96\,s & 5.08\,s & 0.95\,s & 1.37\,s & 3.32\,s & 8.19\,s & 3.31\,s & 4.17\\
WWL & 13.19\,s & 79.33\,s & 11.17\,s & 29.40\,s & 24.10\,s & 90.97\,s & 41.36\,s & 12.00\\
SWWL & 3.51\,s & 5.87\,s & 2.74\,s & 3.35\,s & 4.43\,s & 9.91\,s & 4.97\,s & 5.50\\
NSPDK & 5.78\,s & 4.26\,s & 3.07\,s & 2.81\,s & 2.24\,s & 3.11\,s & 3.54\,s & 4.33\\
NSPDK (disc.) & 5.78\,s & 4.26\,s & 3.07\,s & 2.81\,s & 2.24\,s & 3.11\,s & 3.54\,s & 4.33\\
MMD-GK & 43.03\,s & 150.76\,s & 36.84\,s & 51.87\,s & 79.31\,s & 333.70\,s & 115.92\,s & 14.17\\
\rowcolor{gray!10}
\textbf{NSPPK (ours)} & 11.31\,s & 34.55\,s & 6.09\,s & 6.09\,s & 5.58\,s & 4.72\,s & 11.39\,s & 8.00\\
\bottomrule
\end{tabular}
\end{adjustbox}
\end{table}

\FloatBarrier
\section{Large-Scale Experiment: MolPCBA Learning Curves and Efficiency}
\label{sec:appendix:large_dataset}

The Open Graph Benchmark dataset \texttt{ogbg-molpcba} contains 437{,}929
molecular graphs and 128 binary prediction targets with missing
labels~\citep{hu2020ogb}.

\subsection{Implementation Details for Task~95 Experiments}
\label{appendix:task95}
We evaluate task~95 of \texttt{ogbg-molpcba}, which contains
48{,}853 positive labels, 260{,}115 negative labels, and
128{,}961 missing labels. A balanced cohort of 97{,}706
graphs is constructed by retaining all positive examples
and randomly sampling an equal number of negative examples.
Balanced training subsets ranged from 50 to 78{,}164 examples. The
learning-curve results summarize five runs at each of 13 training sizes. We
report average precision (AP) and ROC--AUC.
\paragraph{NSPPK Configuration.}
For all NSPPK experiments we fixed the parameters across the task:
\[
R=1, \quad D=4, \quad c=1, \quad b=16.
\]
Both sparse and dense representations were evaluated. Sparse mode stores only
nonzero entries, while dense mode expands the full vectors; their measured
training-time differences are reported below.

\paragraph{Downstream Classifiers.}
The exact settings for the classifier paired with NSPPK features are given in Table~\ref{tab:nsppk-classifiers}.

\begin{table}[H]
\centering
\caption{Classifiers used with NSPPK features on \texttt{ogbg-molpcba} (task 95).}
\label{tab:nsppk-classifiers}
\begin{adjustbox}{max width=\textwidth}
\begin{tabular}{lll}
\toprule
Classifier & Features & Configuration \\
\midrule
Logistic Regression & Sparse & \texttt{saga}, max\_iter=1000, $L_2$, $n_{\text{jobs}}=64$ \\
Random Forest       & Sparse & 500 trees, depth=5, $n_{\text{jobs}}=64$, seed=42 \\
XGBoost             & Dense  & 1000 trees, depth=6, LR=0.03, subsample=0.8, colsample=0.8 \\
XGBoost             & Sparse   & 1000 trees, depth=6, LR=0.03, subsample=0.8, colsample=0.8, \texttt{hist} backend \\
\bottomrule
\end{tabular}
\end{adjustbox}
\end{table}

\paragraph{End-to-End Efficiency Accounting.}
Table~\ref{tab:task95_efficiency} separates feature construction and
vectorization from downstream training on task~95. Values are averaged over
training sizes and five random seeds per size.
For kernel methods and NSPPK pipelines, extraction denotes separately
measured feature construction and vectorization time. For the neural
baselines, graph processing is included in training time.

\begin{table}[H]
\centering
\footnotesize
\caption{Mean predictive performance and runtime on \texttt{ogbg-molpcba}
(task~95), averaged over learning-curve training sizes and five random seeds.
Extraction and training times are reported separately when feature construction
is a distinct preprocessing stage.}
\label{tab:task95_efficiency}
\begin{adjustbox}{max width=\textwidth}
\begin{tabular}{lccccc}
\toprule
Method & Mean Test AP & Mean Test AUC & Mean Extraction & Mean Training & Mean Total \\
\midrule
NSPPK-XGBoost-Dense        & 0.755 & 0.770 & 51.0\,s  & 67.2\,s  & 118.2\,s \\
FNP                        & 0.750 & 0.766 & 0.0\,s   & 506.0\,s & 506.0\,s \\
NSPPK-XGBoost-Sparse       & 0.744 & 0.759 & 53.5\,s  & 22.4\,s  & 75.9\,s \\
NSPDK                      & 0.732 & 0.746 & 33.8\,s  & 91.6\,s  & 125.4\,s \\
NSPPK-RandomForest-Sparse  & 0.729 & 0.741 & 53.5\,s  & 10.9\,s  & 64.4\,s \\
NSPPK-LogReg-Sparse        & 0.728 & 0.740 & 53.5\,s  & 550.3\,s & 603.8\,s \\
PDF                        & 0.716 & 0.726 & 0.0\,s   & 887.2\,s & 887.2\,s \\
WL                         & 0.712 & 0.722 & 112.3\,s & 390.0\,s & 502.3\,s \\
SWWL                       & 0.700 & 0.714 & 7.3\,s   & 0.9\,s   & 8.2\,s \\
GNN                        & 0.696 & 0.724 & 0.0\,s   & 298.6\,s & 298.6\,s \\
GAT                        & 0.695 & 0.724 & 0.0\,s   & 316.9\,s & 316.9\,s \\
PNA                        & 0.695 & 0.722 & 0.0\,s   & 468.2\,s & 468.2\,s \\
GIN                        & 0.695 & 0.723 & 0.0\,s   & 214.8\,s & 214.8\,s \\
\bottomrule
\end{tabular}
\end{adjustbox}
\end{table}

SWWL and NSPDK have lower mean extraction times than sparse NSPPK. Among the
NSPPK pairings, sparse random forest has the lowest mean total time
(64.4\,s), and dense XGBoost has the highest mean test AP (0.755).

\paragraph{Neural Baselines.}
Table~\ref{tab:nn-baselines} gives the experimental configurations and
cites the corresponding methods.

\begin{table}[H]
\centering
\caption{Neural and kernel baseline configurations for task 95.}
\label{tab:nn-baselines}
\begin{adjustbox}{max width=\textwidth}
\begin{tabular}{lll}
\toprule
Model & Main hyperparameters & Method reference \\
\midrule
PNA & 2 layers, dim 64$\to$32, batch 64/256, LR=0.001, Adam & \cite{corso2020pna} \\
PDF (Basis-DGL) & 8 layers, dim 384, batch 64/256, LR=5e-4, AdamW & \cite{yang2023bettergraphrepresentationlearning} \\
GAT & 2 layers, 64$\to$32, 4/1 heads, LR=0.001, Adam & \cite{velivckovic2018gat} \\
Attentive FP (FNP) & 4 layers, dim 64, dropout=0.2, LR=0.001, Adam & \cite{xiong2019attentivefp} \\
GIN & 2 layers, dim 64$\to$32, LR=0.001, Adam & \cite{xu2019powerful} \\
GCN (OGB baseline) & 2 layers, dim 64$\to$32, LR=0.001, Adam & \cite{hu2020ogb} \\
WL  & Iterations = 5  & \cite{shervashidze2011weisfeiler} \\
SWWL  & Iterations = 3, Projections = 16, Quantiles = 9 & \cite{carpintero2024swwl} \\
NSPDK  & Radius=1, Distance=4, hash bits $b=16$ & \cite{costa2010fast} \\
\bottomrule
\end{tabular}
\end{adjustbox}
\end{table}

\paragraph{Training and evaluation splits.}
The balanced dataset is partitioned using a stratified
random split into 78{,}164 training, 9{,}771 validation,
and 9{,}771 test graphs. At each training-set size, equal
numbers of positive and negative examples are sampled
from the training partition.

The neural baselines are trained using binary cross-entropy
with logits. Unless otherwise specified, all experiments
are conducted on CPUs.

\FloatBarrier
\section{Additional Sensitivity Results}
\label{appendix:sensitivity-results}

\subsection{Classifier Sensitivity}
\label{sec:classifier_sensitivity}

We hold the attributed NSPPK representation fixed and vary the downstream
classifier. All classifiers use $R=1$, $D=4$, $c=1$, 11 hash bits, and no
degree cutoff. XGBoost has the highest mean accuracy across the six datasets,
with ExtraTrees and Random Forest within 0.32 percentage points. Logistic
regression leads on ENZYMES, and ExtraTrees leads or ties on BZR and
SYNTHETICnew (Table~\ref{tab:classifier_sensitivity}).

\begin{table}[H]
\centering
\caption{Classifier sensitivity for the fixed attributed NSPPK representation.
All rows use $R=1$, $D=4$, $c=1$, $b=11$, and no degree cutoff. Results are
mean test accuracy (\%) $\pm$ standard deviation over stratified 10-fold outer
cross-validation.}
\label{tab:classifier_sensitivity}
\scriptsize
\setlength{\tabcolsep}{2.5pt}
\renewcommand{\arraystretch}{1.08}
\begin{adjustbox}{max width=\textwidth}
\begin{tabular}{lccccccc}
\toprule
\textbf{Classifier} & \textbf{ENZYMES} & \textbf{BZR} & \textbf{PROTEINS\_full} &
\textbf{DHFR} & \textbf{SYNTHETICnew} & \textbf{COX2} & \textbf{Average} \\
\midrule
XGBoost & 51.33 $\pm$ 4.36 & 87.43 $\pm$ 3.07 &
\textbf{76.73 $\pm$ 3.72} & \textbf{83.47 $\pm$ 4.06} &
97.67 $\pm$ 2.25 & \textbf{81.60 $\pm$ 4.48} & \textbf{79.71} \\
ExtraTrees & 52.67 $\pm$ 6.10 & \textbf{87.66 $\pm$ 4.19} &
74.13 $\pm$ 4.09 & 81.48 $\pm$ 3.00 &
\textbf{99.67 $\pm$ 1.05} & 80.94 $\pm$ 4.32 & 79.43 \\
Random Forest & 50.67 $\pm$ 7.79 & 86.65 $\pm$ 4.31 &
75.84 $\pm$ 4.11 & 82.14 $\pm$ 3.75 &
\textbf{99.67 $\pm$ 1.05} & 81.38 $\pm$ 3.28 & 79.39 \\
Logistic Regression & \textbf{54.83 $\pm$ 5.85} & 85.43 $\pm$ 3.37 &
71.61 $\pm$ 4.45 & 80.42 $\pm$ 4.61 &
98.33 $\pm$ 2.36 & 81.38 $\pm$ 2.57 & 78.67 \\
Linear SVM & 52.17 $\pm$ 7.12 & 85.92 $\pm$ 3.32 &
68.55 $\pm$ 5.51 & 78.97 $\pm$ 5.14 &
98.00 $\pm$ 2.81 & 78.18 $\pm$ 4.36 & 76.96 \\
Ridge Classifier & 43.17 $\pm$ 6.40 & 80.98 $\pm$ 7.64 &
57.14 $\pm$ 6.18 & 76.72 $\pm$ 3.09 &
96.00 $\pm$ 3.44 & 79.23 $\pm$ 5.33 & 72.21 \\
\bottomrule
\end{tabular}
\end{adjustbox}
\end{table}

\subsection{Hyperparameter Sensitivity}
\label{sec:sensitivity}

We study how predictive performance varies with the anchor radius $R$,
maximum pairwise distance $D$, and maximum connector radius $R'$.
On the COX2 dataset, we perform a stratified 10-fold cross-validation
experiment, varying one parameter at a time while keeping the other two fixed at
$(R,D,c)=(1,4,1)$, using 11 hash bits and the COX2 degree cutoff $\tau=3$.
Table~\ref{tab:coxsensitivity} reports the average accuracy and standard
deviation. Across all three sweeps, accuracy remains very stable: performance
fluctuates by only about $1$--$2$ percentage points across the entire range.
Within this COX2 experiment, fine-grained tuning of $(R,D,c)$ has little
effect on mean accuracy.

\begin{table}[H]
\centering
\caption{Hyperparameter sensitivity for attributed NSPPK on COX2. Each column
varies one structural parameter while fixing the others at
$(R,D,c)=(1,4,1)$, with $b=11$ and $\tau=3$. Results are mean test accuracy
(\%) $\pm$ standard deviation over stratified 10-fold cross-validation. The
reported distance sweep starts at $D=1$.}
\label{tab:coxsensitivity}
\scriptsize
\setlength{\tabcolsep}{4pt}
\begin{tabular}{cccc}
\toprule
Value & Radius $R$ & Distance $D$ & Connector $c$ \\
\midrule
0 & $81.0 \pm 5.0$ & --- & $81.0 \pm 5.0$ \\
1 & $81.0 \pm 5.0$ & $81.0 \pm 5.0$ & $81.0 \pm 5.0$ \\
2 & $80.0 \pm 3.0$ & $80.0 \pm 4.0$ & $79.0 \pm 3.0$ \\
3 & $80.0 \pm 5.0$ & $80.0 \pm 5.0$ & $80.0 \pm 3.0$ \\     
4 & $80.0 \pm 3.0$ & $81.0 \pm 5.0$ & $80.0 \pm 4.0$ \\
5 & $80.0 \pm 3.0$ & $80.0 \pm 4.0$ & $79.0 \pm 4.0$ \\
6 & $79.0 \pm 3.0$ & $80.0 \pm 4.0$ & $80.0 \pm 4.0$ \\
\bottomrule
\end{tabular}
\end{table}

\subsection{Density Sensitivity}
\label{sec:density_sensitivity}

\begin{table}[!htbp]
  \centering
  \caption{Density sensitivity for NSPPK feature extraction on Erdős--Rényi
  graphs. All rows use $n=300$, $R=1$, $D=4$, $c=1$, and $b=11$. Runtime is
  reported in seconds.}
  \label{tab:density_sensitivity}
  \scriptsize
  \setlength{\tabcolsep}{5pt}
  \begin{tabular}{cc}
  \toprule
  \textbf{Avg.\ Deg.} & \textbf{Time (s)} \\
  \midrule
  2   & 2.29   \\
  27  & 66.11  \\
  52  & 102.40 \\
  77  & 150.75 \\
  102 & 202.39 \\
  127 & 256.25 \\
  152 & 308.38 \\
  177 & 355.59 \\
  202 & 388.98 \\
  227 & 412.80 \\
  252 & 425.54 \\
  277 & 429.48 \\
  \bottomrule
  \end{tabular}
\end{table}
We generated Erdős--Rényi graphs $G(n,q)$ with $n=300$ and
$q=\bar d/(n-1)$ for the target mean degree $\bar d$. All graphs use a
single node label and no continuous attributes. Table~\ref{tab:density_sensitivity}
reports extraction time for $R=1$, $D=4$, $c=1$, and $b=11$.
Runtime rises from 2.29 seconds at target degree 2 to 429.48 seconds at
target degree 277. The increase flattens near the upper end of this
fixed-size sweep.

\end{document}